\documentclass[runningheads]{llncs}
\usepackage{graphicx}
\usepackage{tikz}
\usepackage{comment}
\usepackage{amsmath,amssymb}
\usepackage{color}
\usepackage{multirow}
\usepackage{float}
\usepackage{xcolor}
\usepackage[pagebackref,breaklinks,colorlinks]{hyperref}
\usepackage{cleveref}
\usepackage{svg}
\usepackage{xspace}
\usepackage{subfig}
\usepackage{algorithm}
\usepackage{algpseudocode}
\usepackage{tikz}
\usepackage[numbers,sort,compress]{natbib}

\usepackage[accsupp]{axessibility}
\usepackage[width=122mm,left=12mm,paperwidth=146mm,height=193mm,top=12mm,paperheight=217mm]{geometry}

\usepackage{comment}

\iftrue
	\definecolor{jtcolor}{RGB}{0,0,255}
	\newcommand\JT[1] {\emph{\textcolor{jtcolor}{JT: #1}}}

	\definecolor{todocolor}{RGB}{255,0,00}
	\newcommand\TODO[1] {\PackageWarning{}{Unprocessed todo}\emph{\textcolor{todocolor}{TODO: #1}}}

	\definecolor{wzcolor}{RGB}{0, 200, 0}
	\newcommand\WZ[1] {\emph{\textcolor{wzcolor}{WZ: #1}}}
\else 
	\newcommand\JT[1] {}
	\newcommand\TODO[1] {}
	\newcommand\WZ[2] {}
\fi 

\usepackage{pifont} 
\usepackage{booktabs}
\usepackage{multirow}
\usepackage{adjustbox}
\usepackage{fontawesome} 
\usepackage{amsmath} 
\newcommand{\tableopt}[2]{{\color{#1}\textbf{#2}}}
\definecolor{opA}{rgb}{0.9,0.6,0.0}
\definecolor{opB}{rgb}{0.35,0.70,0.90}
\definecolor{opC}{rgb}{0.8,0.40,0.0}
\definecolor{opD}{rgb}{0.0,0.60,0.50} %
\definecolor{opE}{rgb}{0.8,0.6,0.7}
\definecolor{opF}{rgb}{0.,0.45,0.70} 

\definecolor{pltBlue}{rgb}{0.12156862745098039, 0.4666666666666667, 0.7058823529411765}
\definecolor{pltOrange}{rgb}{1.0, 0.4980392156862745, 0.054901960784313725}
\definecolor{pltGreen}{rgb}{0.17254901960784313, 0.6274509803921569, 0.17254901960784313}
\definecolor{pltRed}{rgb}{0.8392156862745098, 0.15294117647058825, 0.1568627450980392}
\definecolor{pltViolet}{rgb}{0.5803921568627451, 0.403921568627451, 0.7411764705882353}
\definecolor{pltBrown}{rgb}{0.5490196078431373, 0.33725490196078434, 0.29411764705882354}
\definecolor{pltMagenta}{rgb}{0.8901960784313725, 0.4666666666666667, 0.7607843137254902}
\definecolor{pltGray}{rgb}{0.4980392156862745, 0.4980392156862745, 0.4980392156862745}
\definecolor{pltLightGreen}{rgb}{0.7372549019607844, 0.7411764705882353, 0.13333333333333333}
\definecolor{pltCyan}{rgb}{0.09019607843137255, 0.7450980392156863, 0.8117647058823529}
\definecolor{pltPink}{rgb}{0.49803921568627, 0.49803921568627, 0.49803921568627}

\definecolor{cmarkcolor}{rgb}{0.49,0.74,0.49}
\definecolor{xmarkcolor}{rgb}{0.86,0.34,0.34}
\newcommand{\cmark}{\textcolor{cmarkcolor}{\ding{51}}}
\newcommand{\xmark}{\textcolor{xmarkcolor}{\ding{55}}}

\def\link#1{
    \ifx&#1&
        \xmark{}
    \else
        {\href{#1}{\faExternalLink}}
    \fi
}

\newcommand{\tikzcircle}[2][red,fill=red]{\tikz[baseline=-0.5ex]\draw[#1,radius=#2] (0,0) circle ;}%

\newcommand{\model}{MICA\xspace}
\newcommand{\modellong}{\model~(MetrIC fAce)}

\begin{document}
\pagestyle{headings}
\mainmatter
\def\ECCVSubNumber{7474}  
\title{Towards Metrical Reconstruction \\ of Human Faces}
\titlerunning{Towards Metrical Reconstruction of Human Faces}
\author{Wojciech Zielonka \and
Timo Bolkart \and
Justus Thies}

\authorrunning{W. Zielonka et al.}
\institute{Max Planck Institute for Intelligent Systems, Tübingen}
\maketitle

\begin{figure}
    \centering
    \includegraphics[width=\textwidth]{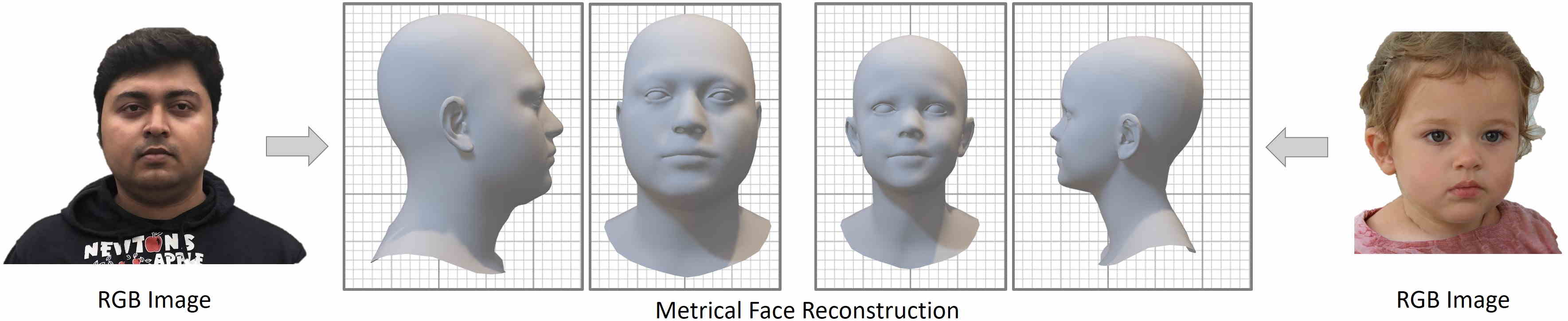}
      \caption{An RGB image of a subject serves as input to \model, which predicts a metrical reconstruction of the human face.
      Images from NoW~\cite{RingNet:CVPR:2019}, StyleGan2~\cite{Karras2019stylegan2}.}
       \label{fig:teaser}
\end{figure}

\begin{abstract}
Face reconstruction and tracking is a building block of numerous applications in AR/VR, human-machine interaction, as well as medical applications.
Most of these applications rely on a metrically correct prediction of the shape, especially, when the reconstructed subject is put into a metrical context (i.e., when there is a reference object of known size).
A metrical reconstruction is also needed for any application that measures distances and dimensions of the subject (e.g., to virtually fit a glasses frame).
State-of-the-art methods for face reconstruction from a single image are trained on large 2D image datasets in a self-supervised fashion.
However, due to the nature of a perspective projection they are not able to reconstruct the actual face dimensions, and even predicting the average human face outperforms some of these methods in a metrical sense.
To learn the actual shape of a face, we argue for a supervised training scheme.
Since there exists no large-scale 3D dataset for this task, we annotated and unified small- and medium-scale databases.
The resulting unified dataset is still a medium-scale dataset with more than 2k identities and training purely on it would lead to overfitting.
To this end, we take advantage of a face recognition network pretrained on a large-scale 2D image dataset, which provides distinct features for different faces and is robust to expression, illumination, and camera changes.
Using these features, we train our face shape estimator in a supervised fashion, inheriting the robustness and generalization of the face recognition network.
Our method, which we call \modellong, outperforms the state-of-the-art reconstruction methods by a large margin, both on current non-metric benchmarks as well as on our metric benchmarks ($15\%$ and $24\%$ lower average error on NoW, respectively).
\\
\textbf{Project website:} \href{https://zielon.github.io/mica/}{https://zielon.github.io/mica/}
\end{abstract}

\section{Introduction}
\label{sec:intro}
Learning to reconstruct 3D content from 2D imagery is an ill-posed inverse problem \cite{BasSmith2019}.
State-of-the-art RGB-based monocular facial reconstruction and tracking methods~\cite{deca,deng2019accurate} are based on self-supervised training, exploiting an underlying metrical face model which is constructed using a large-scale dataset of registered 3D scans (e.g., $33000$ scans for the FLAME~\cite{flame} model).
However, when assuming a perspective camera, the scale of the face is ambiguous since a large face can be modeled by a small face that is close to the camera or a gigantic face that is far away.
Formally, a point $\boldsymbol{x} \in \mathbb{R}^3$ of the face is projected to a point $\boldsymbol{p} \in \mathbb{R}^2$ on the image plane with the projective function $\pi(\cdot)$ and a rigid transformation composed of a rotation $\boldsymbol{R} \in \mathbb{R}^{3\times3}$ and a translation $\boldsymbol{t} \in \mathbb{R}^3$:
$$
\boldsymbol{p} = \pi(\boldsymbol{R} \cdot \boldsymbol{x}+\boldsymbol{t}) = \pi(s \cdot (\boldsymbol{R} \cdot \boldsymbol{x} + \boldsymbol{t})) = \pi(\boldsymbol{R} \cdot (s \cdot \boldsymbol{x}) + (s \cdot \boldsymbol{t}))).
$$
The perspective projection is invariant to the scaling factor $s \in \mathbb{R}$, and thus, if $\boldsymbol{x}$ is scaled by $s$, the rigid transformation can be adapted such that the point still projects onto the same pixel position $\boldsymbol{p}$ by scaling the translation $\boldsymbol{t}$ by $s$.
In consequence, face reconstruction methods might result in a good 2D alignment but can fail to reconstruct the metrical 3D surface and the meaningful metrical location in space.
However, a metric 3D reconstruction is needed in any scenario where the face is put into a metric context.
E.g., when the reconstructed human is inserted into a virtual reality (VR) application or when the reconstructed geometry is used for augmented reality (AR) applications (teleconferencing in AR/VR, virtual try-on, etc.).
In these scenarios, the methods mentioned above fail since they do not reproduce the correct scale and shape of the human face.
In the current literature~\cite{RingNet:CVPR:2019,feng_benchmark,zhu2021facescape}, we also observe that methods use evaluation measurements not done in a metrical space.
Specifically, to compare a reconstructed face to a reference scan, the estimation is aligned to the scan via Procrustes analysis, including an optimal scaling factor.
This scaling factor favors the estimation methods that are not metrical, and the reported numbers in the publications are misleading for real-world applications (relative vs. absolute/metrical error).
In contrast, we aim for a metrically correct reconstruction and evaluation that directly compares the predicted geometry to the reference data without any scaling applied in a post-processing step which is fundamentally different.
As discussed above, the self-supervised methods in the literature do not aim and cannot reconstruct a metrically correct geometry.
However, training these methods in a supervised fashion is not possible because of the lack of data (no large-scale 3D dataset is available).
Training on a small- or medium-scale 3D dataset will lead to overfitting of the networks (see study in the supplemental document).
To this end, we propose a hybrid method that can be trained on a medium-scale 3D dataset, reusing powerful descriptors from a pretrained face recognition network (trained on a large-scale 2D dataset).
Specifically, we propose the usage of existing 3D datasets like LYHM~\cite{lyhm}, FaceWarehouse~\cite{facewarehouse}, Stirling~\cite{stirling}, etc., that contain RGB imagery and corresponding 3D reconstructions to learn a metrical reconstruction of the human head.
To use these 3D datasets, significant work has been invested to unify the 3D data (i.e., to annotate and non-rigidly fit the FLAME model to the different datasets).
This unification provides us with meshes that all share the FLAME topology.
Our method predicts the head geometry in a neutral expression, only given a single RGB image of a human subject in any pose or expression.
To generalize to unseen in the wild images, we use a state-of-the-art face recognition network~\cite{arcface} that provides a feature descriptor for our geometry-estimating network.
This recognition network is robust to head poses, different facial expressions, occlusions, illumination changes, and different focal lengths, thus, being ideal for our task (see \Cref{fig:invariance_qualitative}).
Based on this feature, we predict the geometry of the face with neutral expression within the face space spanned by FLAME~\cite{flame}, effectively disentangling shape and expression.
As an application, we demonstrate that our metrical face reconstruction estimator can be integrated in a new analysis-by-synthesis face tracking framework which removes the requirement of an identity initialization phase~\cite{face2face}.
Given the metrical face shape estimation, the face tracker is able to predict the face motion in a metrical space.

\medskip
\noindent
In summary, we have the following contributions:
\begin{itemize}
    \item a dataset of 3D face reference data for about 2300 subjects, built by unifying existing small- and medium-scale datasets under common FLAME topology.
    \item a metrical face shape predictor -- \model --  which is invariant to expression, pose and illumination, by exploiting generalized identity features from a face recognition network and supervised learning.
    \item a hybrid face tracker that is based on our (learned) metrical reconstruction of the face shape and an optimization-based facial expression tracking.
    \item a metrical evaluation protocol and benchmark, including a discussion on the current evaluation practise.
\end{itemize}

\section{Related Work}
\label{sec:related}
Reconstructing human faces and heads from monocular RGB, RGB-D, or multiview data is a well-explored field at the intersection of computer vision and computer graphics.
Zollhöfer et al.~\cite{Zollhoefer2018FaceSTAR} provide an extensive review of reconstruction methods, focusing on optimization-based techniques that follow the principle of analysis-by-synthesis.
Primarily, the approaches that are based on monocular inputs are based on a prior of face shape and appearance \cite{Garrido2014,Garrido2015,Weise2009,Weise2011,face2face,Thies15,Blanz2003,Blanz2004,deepvideoportraits,deferredneuralrendering,HeadOn,thies2018facevr,thies2020nvp}.
The seminal work of Blanz et al.~\cite{Blanz1999} introduced such a 3D morphable model (3DMM), which represents the shape and appearance of a human in a compressed, low-dimensional, PCA-based space (which can be interpreted as a decoder with a single linear layer).
There is a large corpus of different morphable models~\cite{3DMM_survey}, but the majority of reconstruction methods use either the Basel Face Model~\cite{Blanz1999,bfm09} or the Flame head model~\cite{flame}.
Besides using these models for an analysis-by-synthesis approach, there is a series of learned regression-based methods.
An overview of these methods is given by Morales et al.~\cite{morales2021survey}.
In the following, we will discuss the most relevant related work for monocular RGB-based reconstruction methods.
\paragraph{Optimization-based Reconstruction of Human Faces.}
Along with the introduction of a 3D morphable model for faces, Blanz et al.~\cite{Blanz1999} proposed an optimization-based reconstruction method that is based on the principle of analysis-by-synthesis.
While they used a sparse sampling scheme to optimize the color reproduction, Thies et al.~\cite{Thies15,face2face} introduced a dense color term considering the entire face region that is represented by a morphable model using differentiable rendering.
This method has been adapted for avatar digitization from a single image~\cite{liwen2017} including hair, is used to reconstruct high-fidelity facial reflectance and geometry from a single images~\cite{Yamaguchi2018}, for reconstruction and animation of entire upper bodies~\cite{HeadOn}, or avatars with dynamic textures~\cite{nagano2018}.
Recently, these optimization-based methods are combined with learnable components such as surface offsets or view-dependent surface radiance fields~\cite{grassal2021neural}.
In addition to a photometric reconstruction objective, additional terms based on dense correspondence~\cite{gueler2017densereg} or normal~\cite{Abrevaya_2020_CVPR,grassal2021neural} estimations of neural network can be employed.
Optimization-based methods are also used as a building block for neural rendering methods such as deep video portraits~\cite{deepvideoportraits}, deferred neural rendering~\cite{deferredneuralrendering}, or neural voice puppetry~\cite{thies2020nvp}.
Note that differentiable rendering is not only used in neural rendering frameworks but is also a key component for self-supervised learning of regression-based reconstruction methods covered in the following.
%
\vspace{-0.25cm}
\paragraph{Regression-based Reconstruction of Human Faces.}
%
Learning-based face reconstruction methods can be categorized into supervised and self-supervised approaches.
A series of methods are based on synthetic renderings of human faces to perform a supervised training of a regressor that predicts the parameters of a 3D morphable model~\cite{InverseFaceNet,richardson20163d,dou2017endtoend,richardson2017learning}.
%
Genova et al.~\cite{genova2018unsupervised} propose a 3DMM parameter regression technique that is based on synthetic renderings (where ground truth parameters are available) and real images (where multi-view identity losses are applied). It uses FaceNet~\cite{facenet} to extract features for the 3DMM regression task.
%
Tran et al.~\cite{tran2016regressing} and Chang et al.~\cite{chang2018expnet} (ExpNet) directly regress 3DMM parameters using a CNN trained on fitted 3DMM data.
Tu et al.~\cite{Tu2019} propose a dual training pass for images with and without 3DMM fittings.
%
Jackson et al.~\cite{jackson2017large} propose a model-free approach that reconstructs a voxel-based representation of the human face and is trained on paired 2D image and 3D scan data.
PRN~\cite{feng2018joint} is trained on 'in-the-wild' images with fitted 3DMM reconstructions~\cite{zhu20153ddfa}. It is not restricted to a 3DMM model space and predicts a position map in the UV-space of a template mesh. Instead of working in UV-space, Wei et al.~\cite{wei20193d} propose to use graph convolutions to regress the coordinates of the vertices.
%
%
MoFA~\cite{tewari17MoFA} is a network trained to regress the 3DMM parameters in a self-supervised fashion. As a supervision signal, it uses the dense photometric losses of Face2Face~\cite{face2face}.
Within this framework, Tewari et al. proposed to refine the identity shape and appearance~\cite{tewari2017self} as well as the expression basis~\cite{tewari2019fml} of a linear 3DMM.
In a similar setup, one can also train a non-linear 3DMM~\cite{tran2019towards} or personalized models~\cite{chaudhuri2020personalized}.
RingNet~\cite{RingNet:CVPR:2019} regresses 3DMM parameters and is trained on 2D images using losses on the reproduction of 2D landmarks and shape consistency (different images of the same subject) and shape inconsistency (images of different subjects) losses.
%
%
%
DECA~\cite{deca} extends RingNet with expression dependent offset predictions in UV space. It uses dense photometric losses to train the 3DMM parameter regression and the offset prediction network.
This separation of a coarse 3DMM model and a detailed bump map has been introduced by Tran et al.~\cite{tran2017extreme}.
Chen et al.~\cite{chen2019photo} use a hybrid training composed of self-supervised and supervised training based on renderings to predict texture and displacement maps.
Deng et al.~\cite{deng2019accurate} train a 3DMM parameter regressor based on multi-image consistency losses and 'hybrid-level' losses (photometric reconstruction loss with skin attention masks, and a perception-level loss based on FaceNet~\cite{facenet}).
On the NoW challenge~\cite{RingNet:CVPR:2019}, DECA~\cite{deca} and the method of Deng et al.~\cite{deng2019accurate} show on-par state-of-the-art results.
%
%
Similar to DECA's offset prediction, there are GAN-based methods that predict detailed color maps~\cite{Gecer_2019_CVPR,gecer2021fast} or skin properties~\cite{saito2016photorealistic,Yamaguchi2018,lattas2020avatarme,lattas2021avatarme++} (e.g., albedo, reflectance, normals) in UV-space of a 3DMM-based face reconstruction.
In contrast to these methods, we are interested in reconstructing a metrical 3D representation of a human face and not fine-scale details.
Self-supervised methods suffer from the depth-scale ambiguity (the face scale, translation away from the camera, and the perspective projection are ambiguous) and, thus, predict a wrongly scaled face, even though 3DMM models are by construction in a metrical space.
We rely on a strong supervision signal to learn the metrical reconstruction of a face using high-quality 3D scan datasets which we unified.
In combination with an identity encoder~\cite{arcface} trained on in-the-wild 2D data, including occlusions, different illumination, poses, and expressions, we achieve robust geometry estimations that significantly outperform state-of-the-art methods.

\section{Metrical Face Shape Prediction}
\label{sec:main}
\begin{figure*}
    \includegraphics[width=\textwidth]{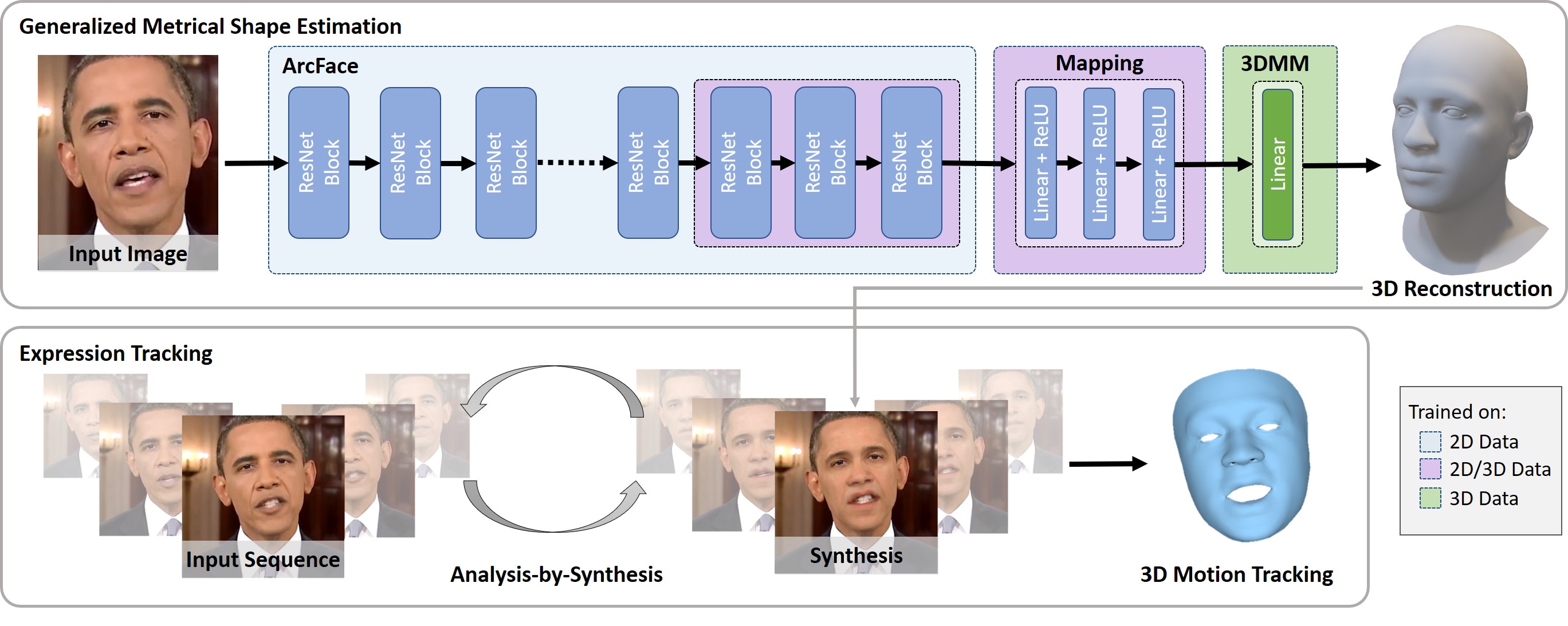}
    \caption{
    We propose a method for metrical human face shape estimation from a single image which exploits a supervised training scheme based on a mixture of different 2D,2D/3D and 3D datasets.
    This estimation can be used for facial expression tracking using analysis-by-synthesis which optimizes for the camera intrinsics, as well as the per-frame illumination, facial expression and pose.
    }
    \label{fig:pipeline}
\end{figure*}

Based on a single input RGB image $I$, \model aims to predict a metrical shape of a human face in a neutral expression.
To this end, we leverage both 'in-the-wild' 2D data as well as metric 3D data to train a deep neural network, as shown in \Cref{fig:pipeline}.
We employ a state-of-the-art face recognition network~\cite{arcface} which is trained on 'in-the-wild' data to achieve a robust prediction of an identity code, which is interpreted by a geometry decoder.

\paragraph{Identity Encoder.}
As an identity encoder, we leverage the ArcFace~\cite{arcface} architecture which is pretrained on Glint360K \cite{an2020partical_fc}.
This ResNet100-based network is trained on 2D image data using an additive angular margin loss to obtain highly discriminative features for face recognition.
It is invariant to illumination, expression, rotation, occlusion, and camera parameters which is ideal for a robust shape prediction.
We extend the ArcFace architecture by a small mapping network $\mathcal{M}$ that maps the ArcFace features to our latent space, which can then be interpreted by our geometry decoder:
\begin{equation*}
    \boldsymbol{z} = \mathcal{M}(ArcFace(I)),
\end{equation*}
where $\boldsymbol{z} \in \mathbb{R}^{300}$. Our mapping network $\mathcal{M}$ consists of three fully-connected linear hidden layers with ReLU activation and the final linear output layer.
\paragraph{Geometry Decoder.}
There are essentially two types of geometry decoders used in the literature, model-free and model-based.
Throughout the project of this paper, we conducted experiments on both types and found that both perform similarly on the evaluation benchmarks.
Since a 3DMM model efficiently represents the face space, we focus on a model-based decoder.
Specifically, we use FLAME~\cite{flame} as a geometry decoder, which consists of a single linear layer:
\begin{equation*}
    \mathcal{G}_{3DMM}(\boldsymbol{z}) = \boldsymbol{B} \cdot \boldsymbol{z} + \boldsymbol{A},
\end{equation*}
where $\boldsymbol{A} \in \mathbb{R}^{3N}$ is the geometry of the average human face and $\boldsymbol{B} \in \mathbb{R}^{3N\times300}$ contains the principal components of the 3DMM and $N=5023$.

\paragraph{Supervised Learning.}
The networks described above are trained using paired 2D/3D data from existing, unified datasets $\mathcal{D}$ (see \Cref{sec:data}).
We fix large portions of the pre-trained ArcFace network during the training and refine the last $3$ ResNet blocks.
Note that ArcFace is trained on a much larger amount of identities, therefore, refining more hidden layers results in worse predictions due to overfitting. We found that using the last $3$ ResNet blocks gives the best generalization (see supplemental document).
The training loss is:
\begin{equation}
    \mathcal{L} = \sum_{(I,\mathcal{G})\in \mathcal{D}} 
    |\kappa_{mask}(\mathcal{G}_{3DMM}(\mathcal{M}(ArcFace(I))) - \mathcal{G})|,
\end{equation}
where $\mathcal{G}$ is the ground truth mesh and $\kappa_{mask}$ is a region dependent weight (the face region has weight $150.0$, the back of the head $1.0$, and eyes with ears $0.01$).
We use AdamW~\cite{adamw} for optimization with fixed learning rate $\eta=1\mathrm{e}{-5}$ and weight decay $\lambda=2\mathrm{e}{-4}$.
We select the best performing model based on the validation set loss using the Florence dataset~\cite{florence}.
The model was trained for $160k$ steps on Nvidia Tesla V100.
\section{Face Tracking}
Based on our shape estimate, we demonstrate optimization-based face tracking on monocular RGB input sequences.
To model the non-rigid deformations of the face, we use the linear expression basis vectors and the linear blendskinning of the FLAME~\cite{flame} model, and use a linear albedo model~\cite{flametexture} to reproduce the appearance of a subject in conjunction with a Lambertian material assumption and a light model based on spherical harmonics.
We adapt the analysis-by-synthesis scheme of Thies et al.~\cite{face2face}.
Instead of using a multi-frame model-based bundling technique to estimate the identity of a subject, we use our one-shot shape identity predictor.
We initialize the albedo and spherical harmonics based on the same first frame using the energy:
\begin{equation}
E(\boldsymbol{\phi}) = w_{dense}E_{dense}(\boldsymbol{\phi}) + w_{lmk}E_{lmk}(\boldsymbol{\phi}) + w_{reg}E_{reg}(\boldsymbol{\phi}),
\label{eq:objective_function}
\end{equation}
where $\boldsymbol{\phi}$ is the vector of unknown parameters we are optimizing for. The energy terms $E_{dense}(\boldsymbol{\phi})$ and $E_{reg}(\boldsymbol{\phi})$ measure the dense color reproduction of the face ($\ell_1$-norm) and the deviation from the neutral pose respectively.
The sparse landmark term $E_{lmk}(\boldsymbol{\phi})$ measures the reproduction of 2D landmark positions (based on Google's mediapipe~\cite{kartynnik2019realtime,grishchenko2020attention} and Face Alignment \cite{bulat2017far}).
The weights $w_{dense}$, $ w_{lmk}$ and $w_{reg}$ balance the influence of each sub-objectives on the final loss.
In the first frame vector $\boldsymbol{\phi}$ contains the 3DMM parameters for albedo, expression, and rigid pose, as well as the spherical harmonic coefficients (3 bands) that are used to represent the environmental illumination \cite{envmap}.
After initialization, the albedo parameters are fixed and unchanged throughout the sequence tracking.

\paragraph{Optimization.}
We optimize the objective function \Cref{eq:objective_function} using Adam~\cite{adam} in PyTorch.
While recent soft-rasterizers~\cite{liu2019softras,ravi2020pytorch3d} are popular, we rely on a sampling based scheme as introduced by Thies et al.~\cite{face2face} to implement the differentiable rendering for the photo-metric reproduction error $E_{dense}(\boldsymbol{\phi})$.
Specifically, we use a classical rasterizer to render the surface of the current estimation.
The rasterized surface points that survive the depth test are considered as the set of visible surface points $\mathcal{V}$ for which we compute the energy term $E_{dense}(\boldsymbol{\phi}) = \sum_{i \in \mathcal{V}} |I(\pi(\boldsymbol{R} \cdot p_i(\boldsymbol{\phi}) + \boldsymbol{t})) - c_i(\boldsymbol{\phi})|$ where $p_i$ and $c_i$ being the $i$-th vertex and color of the reconstructed model, and $I$ the RGB input image.

\section{Dataset Unification}
\label{sec:data}
In the past, methods and their training scheme were limited by the availability of 3D scan datasets of human faces.
While several small and medium-scale datasets are available, they are in different formats and do not share the same topology.
To this end, we unified the available datasets such that they can be used as a supervision signal for face reconstruction from 2D images.
Specifically, we register the FLAME~\cite{flame} head model to the provided scan data.
In an initial step, we fit the model to landmarks and optimize for the FLAME parameters based on an iterative closest point (ICP) scheme~\cite{besl1992icp}.
We further jointly optimize FLAME's model parameters, and refine the fitting with a non-rigid deformation regularized by FLAME, similar to Li and Bolkart et al.~\cite{flame}.
In \Cref{tab:datasets_overview}, we list the datasets that we unified for this project.
We note that the datasets vary in the capturing modality and capturing script (with and without facial expressions, with and without hair caps, indoor and outdoor imagery, still images, and videos), which is suitable for generalization.
The datasets are recorded in different regions of the world and are often biased towards ethnicity. Thus, combining other datasets results in a more diverse data pool.
In the supplemental document, we show an ablation on the different datasets.
\textit{
Upon agreement of the different dataset owners, we will share our unified dataset, i.e., for each subject one registered mesh with neutral expression in FLAME topology.
}
Note that in addition to the datasets listed in \Cref{tab:datasets_overview}, we analyzed the FaceScape dataset~\cite{yang2020facescape}.
While it provides a large set of 3D reconstructions ($\sim 17$k), which would be ideal for our training, the reconstructions are not done in a metrical space.
Specifically, the data has been captured in an uncalibrated setup and faces are normalized by the eye distance, which has not been detailed in their paper (instead, they mention sub-millimeter reconstruction accuracy which is not valid).
This is a fundamental flaw of this dataset, and also questions their reconstruction benchmark~\cite{zhu2021facescape}.

\def\kinect{\tableopt{opA}{K}}
\def\passivestereo{\tableopt{opB}{S}}

\setlength{\tabcolsep}{4pt}
\begin{table}[t!]
    \centering
        \caption{
        Overview of our unified datasets.
        The used datasets vary in the capture modality and the capture protocol.
        Here, we list the number of subject, the minimum number of images per subjects, and whether the dataset includes facial expressions.
        In total our dataset contains 2315 subjects with FLAME topology.
    }
    \medskip
    \begin{tabular}{lcrrc}
        \toprule 
        Dataset & & \#Subj. & \#Min. Img. & Expr.\\
        \midrule
        Stirling~\cite{stirling} & \link{http://pics.stir.ac.uk/ESRC/} & 
        $133$ & $8$ & \cmark{}\\
        D3DFACS~\cite{D3DFACS} & \link{https://www.cs.bath.ac.uk/~dpc/D3DFACS/} & 
        $10$ & videos & \cmark{}\\
        Florence 2D/3D~\cite{florence} & \link{https://www.micc.unifi.it/resources/datasets/florence-3d-faces/} & 
        $53$ & videos & \cmark{}\\
        BU-3DFE~\cite{bu3dfe} & \link{https://www.cs.binghamton.edu/~lijun/Research/3DFE/3DFE_Analysis.html} & 
        $100$ & $83$ & \cmark{}\\
        LYHM~\cite{lyhm} & \link{https://www-users.cs.york.ac.uk/~nep/research/LYHM/} & 
        $1211$ & $2$ & \xmark{}\\
        FaceWarehouse~\cite{facewarehouse} & \link{http://kunzhou.net/zjugaps/facewarehouse/} & $150$ & $119$ & \cmark{}\\
        FRGC~\cite{frgc} & \link{https://cvrl.nd.edu/projects/data/\#face-recognition-grand-challenge-frgc-v20-data-collection} & $531$ & $7$ & \cmark{}\\
        BP4D+~\cite{pb4d} & \link{http://www.cs.binghamton.edu/~lijun/Research/3DFE/3DFE_Analysis.html} & $127$ & videos & \cmark{}\\
        \bottomrule 
    \end{tabular}

    \label{tab:datasets_overview}
\end{table}

\begin{table*}[ht!]
    \caption{
         Quantitative evaluation of the face shape estimation on the \textit{NoW Challenge}~\cite{RingNet:CVPR:2019}. Note that we list two different evaluations: the non-metrical evaluation from the original NoW challenge and our new metrical evaluation (including a cumulative error plot on the left).
        The original NoW challenge cannot be considered metrical since Procrustes analysis is used to align the reconstructions to the corresponding reference meshes, including scaling.
        We list all methods from the original benchmark and additionally show the performance of the average human face of FLAME~\cite{flame} as a reference (first row).
    }
    \medskip
    \centering
    \resizebox{\linewidth}{!}{%
    \begin{tabular}{ll|ccc|ccc}
        \toprule 
        \textbf{NoW-Metric Challenge} & & \multicolumn{3}{c|}{Non-Metrical~\cite{RingNet:CVPR:2019}} & \multicolumn{3}{c}{Metrical (mm)}\\
        \toprule 
        \multirow{7}{*}[-0.7cm]{\hspace{-0.1cm}\vspace{0.5cm}
        \includegraphics[width=6cm]{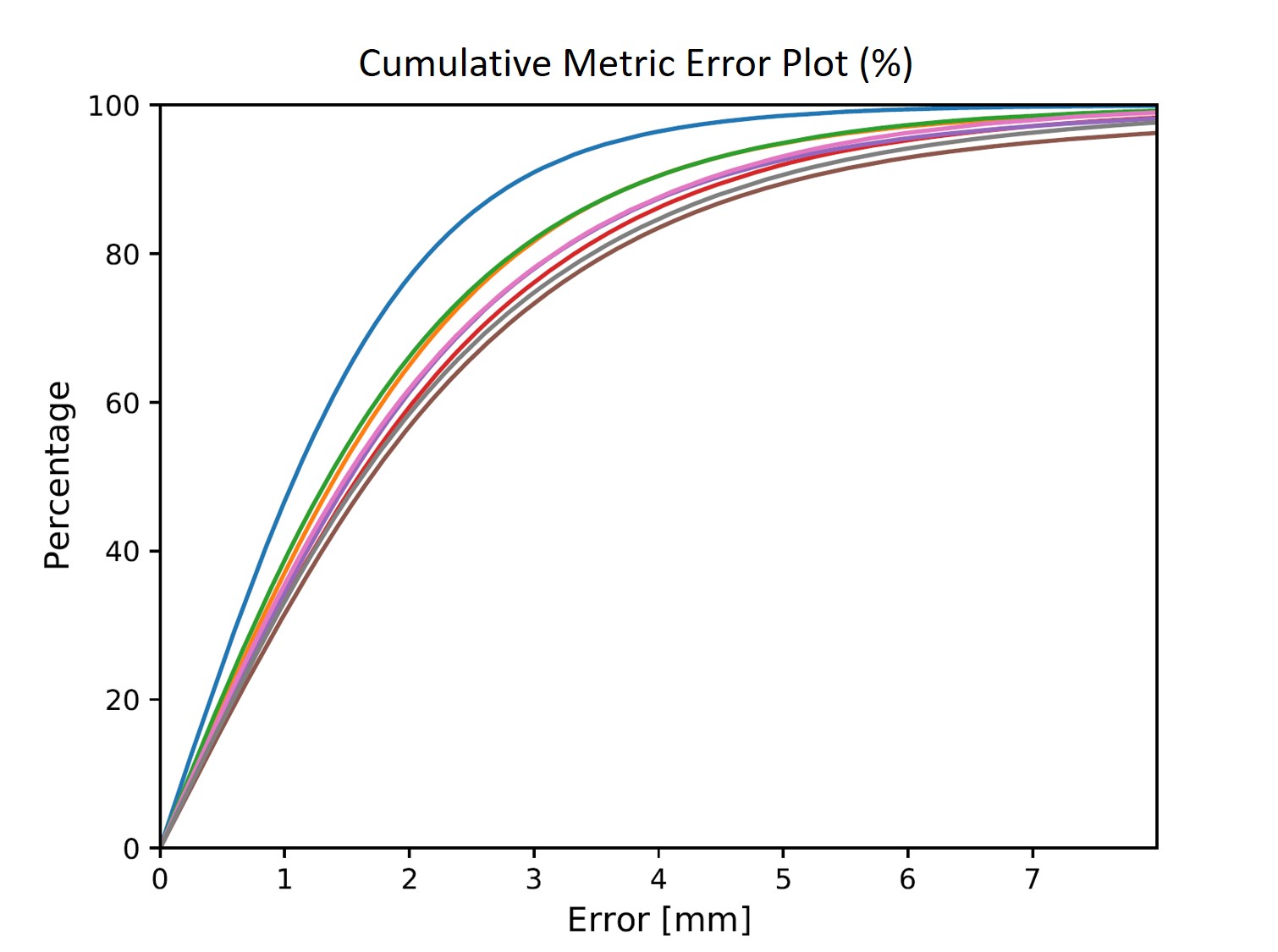}
        }
        & Method & Median & Mean & Std & Median & Mean & Std \\
        \cmidrule{2-8}
        & \tikzcircle[white, fill=white]{1.5pt} Average Face (FLAME~\cite{flame})
        & $1.21$ & $1.53$ & $1.31$
        & $1.49$ & $1.92$ & $1.68$
        \\          
        & \tikzcircle[white, fill=white]{1.5pt} 3DMM-CNN~\cite{tran2016regressing}
        & $1.84$ & $2.33$ & $2.05$
        & $3.91$ & $4.84$ & $4.02$
        \\          
        & \tikzcircle[white, fill=white]{1.5pt} PRNet~\cite{feng2018joint}
        %
        & $1.50$ & $1.98$ & $1.88$
        & -- & -- & --
        \\            
        & \tikzcircle[white, fill=white]{1.5pt} Deng et al~\cite{deng2019accurate} (TensorFlow)
        & $1.23$ & $1.54$ & $1.29$
        & $2.26$ & $2.90$ & $2.51$
        \\
        & \tikzcircle[pltPink, fill=pltPink]{1.5pt} Deng et al~\cite{deng2019accurate} (PyTorch)
        & $1.11$ & $1.41$ & $1.21$
        & $1.62$ & $2.21$ & $2.08$
        \\
        & \tikzcircle[pltMagenta, fill=pltMagenta]{1.5pt} RingNet~\cite{RingNet:CVPR:2019}
        & $1.21$ & $1.53$ & $1.31$
        & $1.50$ & $1.98$ & $1.77$
        \\
        & \tikzcircle[pltBrown, fill=pltBrown]{1.5pt} 3DDFA-V2~\cite{guo2020towards}
        & $1.23$ & $1.57$ & $1.39$
        & $1.53$ & $2.06$ & $1.95$
        \\
        & \tikzcircle[pltRed, fill=pltRed]{1.5pt} MGCNet~\cite{Shang:ECCV:2020}
        & $1.31$ & $1.87$ & $2.63$
        & $1.70$ & $2.47$ & $3.02$
        \\  
        & \tikzcircle[white, fill=white]{1.5pt} UMDFA~\cite{Koizumi:ECCV:2020}
        & $1.52$ & $1.89$ & $1.57$
        & $2.31$ & $2.97$ & $2.57$
        \\          
        & \tikzcircle[pltViolet, fill=pltViolet]{1.5pt} Dib et al.~\cite{Dib:ICCV:2021}
        & $1.26$ & $1.57$ & $1.31$
        & $1.59$ & $2.12$ & $1.93$
        \\          
        & \tikzcircle[pltGreen, fill=pltGreen]{1.5pt} DECA~\cite{deca}
        & $1.09$ & $1.38$ & $1.18$
        & $1.35$ & $1.80$ & $1.64$
        \\
        & \tikzcircle[pltOrange, fill=pltOrange]{1.5pt} FOCUS~\cite{focus}
        & $1.04$ & $1.30$ & $1.10$
        & $1.41$ & $1.85$ & $1.70$
        \\
        & \tikzcircle[pltBlue, fill=pltBlue]{1.5pt} \textbf{Ours} 
        & \textbf{0.90} & \textbf{1.11} & \textbf{0.92}
        & \textbf{1.08} & \textbf{1.37} & \textbf{1.17}
        \\
        \bottomrule 
    \end{tabular}
    }

    \label{tab:now_challenge}
\end{table*}

\section{Results}
\label{sec:results}

Our experiments mainly focus on the metrical reconstruction of a human face from in the wild images.
In the supplemental document, we show results for the sequential tracking of facial motions using our metrical reconstruction as initialization.
The following experiments are conducted with the original models of the respective publications including their reconstructions submitted to the given benchmarks.
Note that these models are trained on their large-scale datasets, training them on our medium-scale 3D dataset would lead to overfitting.
%

\subsection{Face Shape Estimation}
In recent publications, face shape estimation is evaluated on datasets where reference scans of the subjects are available.
The NoW Challenge~\cite{RingNet:CVPR:2019} and the benchmark of Feng et al.~\cite{feng_benchmark} which is based on Stirling meshes~\cite{stirling} are used in the state-of-the-art methods \cite{deca,deng2019accurate,RingNet:CVPR:2019}.
We conduct several studies on these benchmarks and propose different evaluation protocols.

\medskip \noindent
\textbf{Non-Metrical Benchmark.}
The established evaluation methods on these datasets are based on an optimal scaling step, i.e., to align the estimation to the reference scan, they optimize for a rigid alignment and an additional scaling factor which results in a non-metric/relative error.
This scaling compensates for shape mispredictions, e.g., the mean error evaluated on the NoW Challenge for the average FLAME mesh (Table \ref{tab:now_challenge}) drops from 1.92mm to 1.53mm because of the applied scale optimization.
This is an improvement of around $20\%$ which has nothing to do with the reconstruction quality and, thus, creates a misleading benchmark score where methods appear better than they are.
Nevertheless, we evaluate our method on these benchmarks and significantly outperform all state-of-the-art methods as can be seen in Tables \ref{tab:now_challenge} and \ref{tab:feng_challenge} (`Non-Metrical' column).

\begin{table*}[h!]
        \caption{
        Quantitative evaluation of the face shape estimation on the \textit{Stirling Reconstruction Benchmark}~\cite{feng_benchmark} using the NoW protocol~\cite{RingNet:CVPR:2019}.
        We list two different evaluations: the non-metric evaluation from the original benchmark and the metric evaluation.
        \textit{Note that for this experiment, we exclude the Stirling dataset from our training set.}
    }
    \medskip
    \centering
    \resizebox{\linewidth}{!}{%
    \begin{tabular}{ l|*{11}{c|}c }
    \toprule 
    \multirow{3}{*}{\textbf{Stirling  (NoW Protocol)}} 
    & \multicolumn{6}{c|}{Non-Metrical} 
    & \multicolumn{6}{c}{Metrical (mm)}\\
    \cline{2-13}
    & \multicolumn{2}{c|}{Median}
    & \multicolumn{2}{c|}{Mean}
    & \multicolumn{2}{c|}{Std}
    & \multicolumn{2}{c|}{Median}                
    & \multicolumn{2}{c|}{Mean}          
    & \multicolumn{2}{c}{Std}          
    \\
    \cline{2-13}
      & LQ & HQ & LQ & HQ & LQ & HQ & LQ & HQ & LQ & HQ & LQ & HQ \\
    \toprule
    Average Face (FLAME~\cite{flame})
    & $1.23$ & $1.22$ & $1.56$ & $1.55$ & $1.38$ & $1.35$
    & $1.44$ & $1.40$ & $1.84$ & $1.79$ & $1.64$ & $1.57$
    \\
    RingNet~\cite{RingNet:CVPR:2019}
    & $1.17$ & $1.15$ & $1.49$ & $1.46$ & $1.31$ & $1.27$
    & $1.37$ & $1.33$ & $1.77$ & $1.72$ & $1.60$ & $1.54$
    \\
    3DDFA-V2~\cite{guo2020towards}
    & $1.26$ & $1.20$ & $1.63$ & $1.55$ & $1.52$ & $1.45$
    & $1.49$ & $1.38$ & $1.93$ & $1.80$ & $1.78$ & $1.68$
    \\
    Deng et al.~\cite{deng2019accurate} (TensorFlow)
    & $1.22$ & $1.13$ & $1.57$ & $1.43$ & $1.40$ & $1.25$
    & $1.85$ & $1.81$ & $2.41$ & $2.29$ & $2.16$ & $1.97$
    \\
    Deng et al.~\cite{deng2019accurate} (PyTorch)
    & $1.12$ & $0.99$ & $1.44$ & $1.27$ & $1.31$ & $1.15$
    & $1.47$ & $1.31$ & $1.93$ & $1.71$ & $1.77$ & $1.57$
    \\
    DECA~\cite{deca}
    & $1.09$ & $1.03$ & $1.39$ & $1.32$ & $1.26$ & $1.18$
    & $1.32$ & $1.22$ & $1.71$ & $1.58$ & $1.54$ & $1.42$
    \\
    \textbf{Ours w/o. Stirling}
    & \textbf{0.96} & \textbf{0.92} & \textbf{1.22} & \textbf{1.16} & \textbf{1.11} & \textbf{1.04}
    & \textbf{1.15} & \textbf{1.06} & \textbf{1.46} & \textbf{1.35} & \textbf{1.30} & \textbf{1.20}
    \\
    \bottomrule
    \end{tabular}
    }

    \label{tab:feng_challenge_now}
\end{table*}

\begin{table*}[h!]
        \caption{
        Quantitative evaluation of the face shape estimation on the \textit{Stirling Reconstruction Benchmark}~\cite{feng_benchmark}. 
        We list two different evaluations: the non-metric evaluation from the original benchmark and the metric evaluation.
        This benchmark is based on an alignment protocol that only relies on reference landmarks and, thus, is very noisy and dependent on the landmark reference selection (in our evaluation, we use the landmark correspondences provided by the FLAME~\cite{flame} model).
        We use the image file list from \cite{RingNet:CVPR:2019} to compute the scores (i.e., excluding images where a face is not detectable).
        \textit{Note that for this experiment, we exclude the Stirling dataset from our training set.}
    }
    \medskip
    \centering
    \resizebox{\linewidth}{!}{%
    \begin{tabular}{l|*{11}{c|}c }
    \toprule 
    \multirow{3}{*}{\textbf{Stirling/ESRC Benchmark}} 
    & \multicolumn{6}{c|}{Non-Metrical~\cite{feng_benchmark}} 
    & \multicolumn{6}{c}{Metrical (mm)}\\
    \cline{2-13}
    & \multicolumn{2}{c|}{Median}
    & \multicolumn{2}{c|}{Mean}
    & \multicolumn{2}{c|}{Std}
    & \multicolumn{2}{c|}{Median}                
    & \multicolumn{2}{c|}{Mean}          
    & \multicolumn{2}{c}{Std}          
    \\
    \cline{2-13}
      & LQ & HQ & LQ & HQ & LQ & HQ & LQ & HQ & LQ & HQ & LQ & HQ \\
    \toprule
     Average Face (FLAME~\cite{flame})
    & $1.58$ & $1.62$ & $2.06$ & $2.08$ & $1.82$ & $1.83$
    & $1.70$ & $1.62$ & $2.19$ & $2.09$ & $1.96$ & $1.85$
    \\
    RingNet~\cite{RingNet:CVPR:2019}
    & $1.56$ & $1.60$ & $2.01$ & $2.05$ & $1.75$ & $1.76$
    & $1.67$ & $1.64$ & $2.16$ & $2.09$ & $1.90$ & $1.81$
    \\
    3DDFA-V2~\cite{guo2020towards}
    & $1.58$ & $1.49$ & $2.03$ & $1.90$ & $1.74$ & $1.63$
    & $1.70$ & $1.56$ & $2.16$ & $1.98$ & $1.88$ & $1.70$
    \\
    Deng et al.~\cite{deng2019accurate} (TensorFlow)
    & $1.56$ & $1.41$ & $2.02$ & $1.84$ & $1.77$ & $1.63$
    & $2.13$ & $2.14$ & $2.71$ & $2.65$ & $2.33$ & $2.12$
    \\
    Deng et al.~\cite{deng2019accurate} (PyTorch)
    & $1.51$ & $1.29$ & $1.95$ & $1.64$ & $1.71$ & $1.39$
    & $1.78$ & $1.54$ & $2.28$ & $1.97$ & $1.97$ & $1.68$
    \\
    DECA~\cite{deca}
    & $1.40$ & $1.32$ & $1.81$ & $1.72$ & $1.59$ & $1.50$
    & $1.56$ & $1.45$ & $2.03$ & $1.87$ & $1.81$ & $1.64$
    \\
    \textbf{Ours w/o. Stirling}
    & \textbf{1.26} & \textbf{1.22} & \textbf{1.62} & \textbf{1.55} & \textbf{1.41} & \textbf{1.34}
    & \textbf{1.36} & \textbf{1.26} & \textbf{1.73} & \textbf{1.60} & \textbf{1.48} & \textbf{1.37}
    \\
    \hline
    \end{tabular}
    }

    \label{tab:feng_challenge}
\end{table*}

\begin{figure}[ht!]
    \centering
    \includegraphics[width=0.95\linewidth]{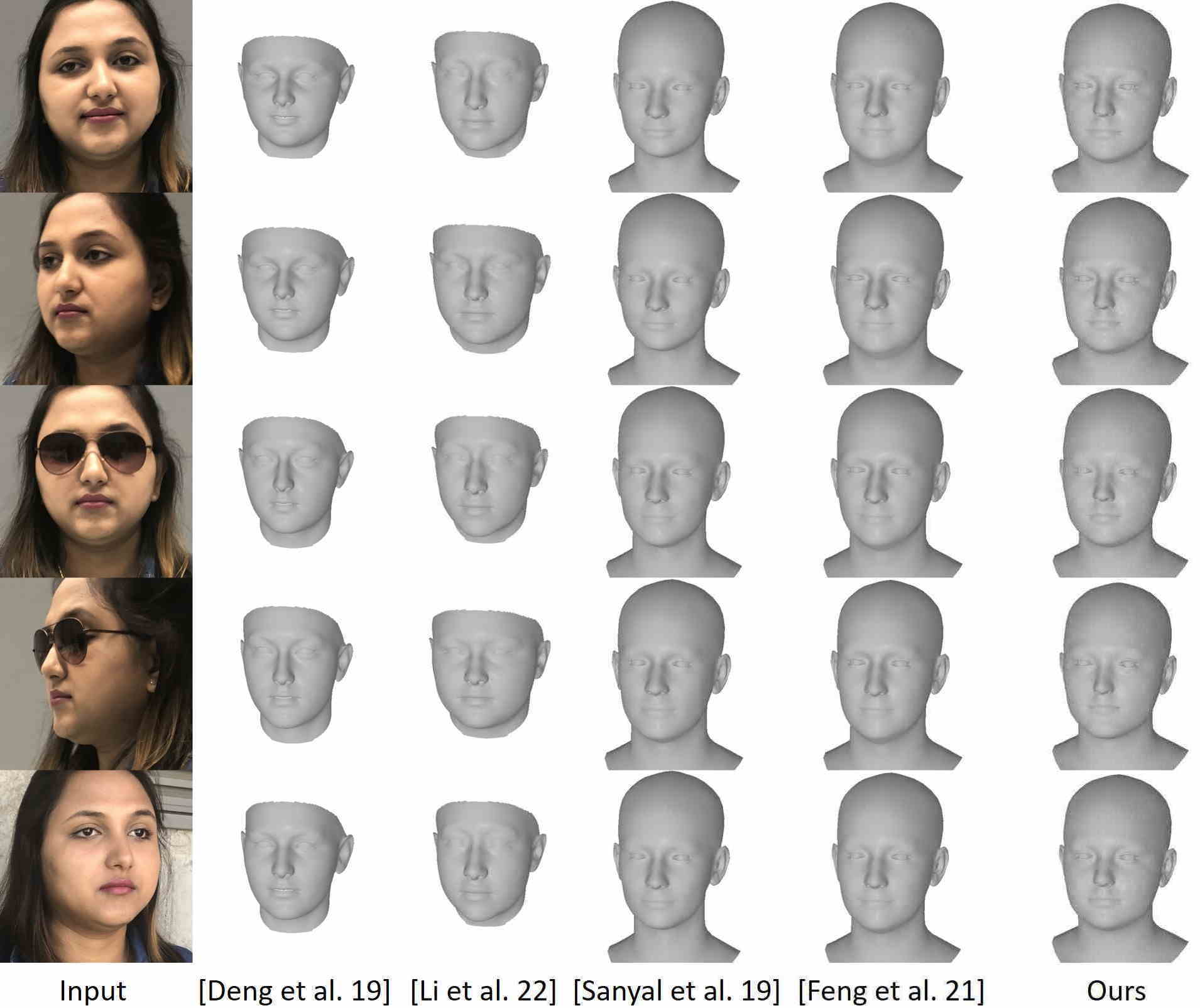}
    \caption[ArcFace Invariance]{Qualitative results on NoW Challenge~\cite{RingNet:CVPR:2019} to show the invariance of our method to changes in illumination, expression, occlusion, rotation, and  perspective distortion in comparison to other methods.}
    \label{fig:invariance_qualitative}
\end{figure}

\begin{figure*}[h!]
    \centering
    \includegraphics[width=\linewidth]{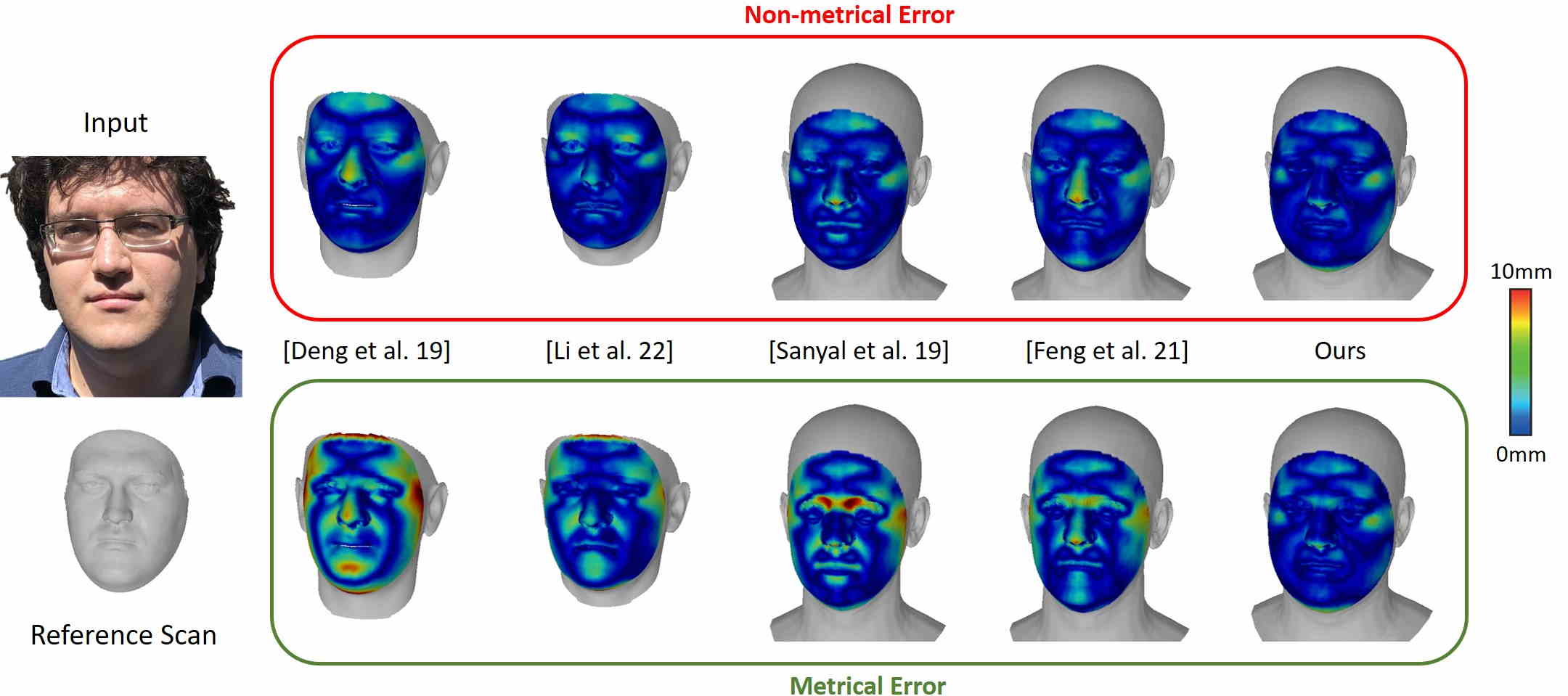}
    \caption[Qualitative Results NoW and NoW-Metric]{
    Established evaluation benchmarks like \cite{feng_benchmark,RingNet:CVPR:2019} are based on a non-metrical error metric (top-row).
    We propose a new evaluation protocol which measures reconstruction errors in a metrical space (bottom row) (c.f. \Cref{tab:now_challenge}).
    Image from the NoW~\cite{RingNet:CVPR:2019} validation set.
    }
    \label{fig:qualitative_now_heatmap_all}
\end{figure*}

\medskip \noindent
\textbf{Metrical Benchmark.}
Since for a variety of applications, actual metrical reconstructions are required, we argue for a new evaluation scheme that uses a purely rigid alignment, i.e., without scale optimization (see \Cref{fig:qualitative_now_heatmap_all}). The error is calculated using an Euclidean distance between each scan vertex and the closest point on the mesh surface.
This new evaluation scheme enables a comparison of methods based on metrical quantities (see Tables \ref{tab:now_challenge} and \ref{tab:feng_challenge}) and, thus, is \textit{fundamentally} different from the previous evaluation schemes.
In addition, the benchmark of Feng et al.~\cite{feng_benchmark} is based on the alignment using sparse facial (hand-selected) landmarks.
Our experiments showed that this scheme is highly dependent on the selection of these markers and results in inconsistent evaluation results.
In our listed results, we use the marker correspondences that come with the FLAME model~\cite{flame}.
To get a more reliable evaluation scheme, we evaluate the benchmark of Feng et al. using the dense iterative closest point (ICP) technique from the NoW challenge, see \Cref{tab:feng_challenge_now}.
On all metrics, our proposed method significantly improves the reconstruction accuracy.
Note that some methods are even performing worse than the mean face~\cite{flame}.

\begin{figure}[ht!]
    \centering
    \includegraphics[width=\linewidth]{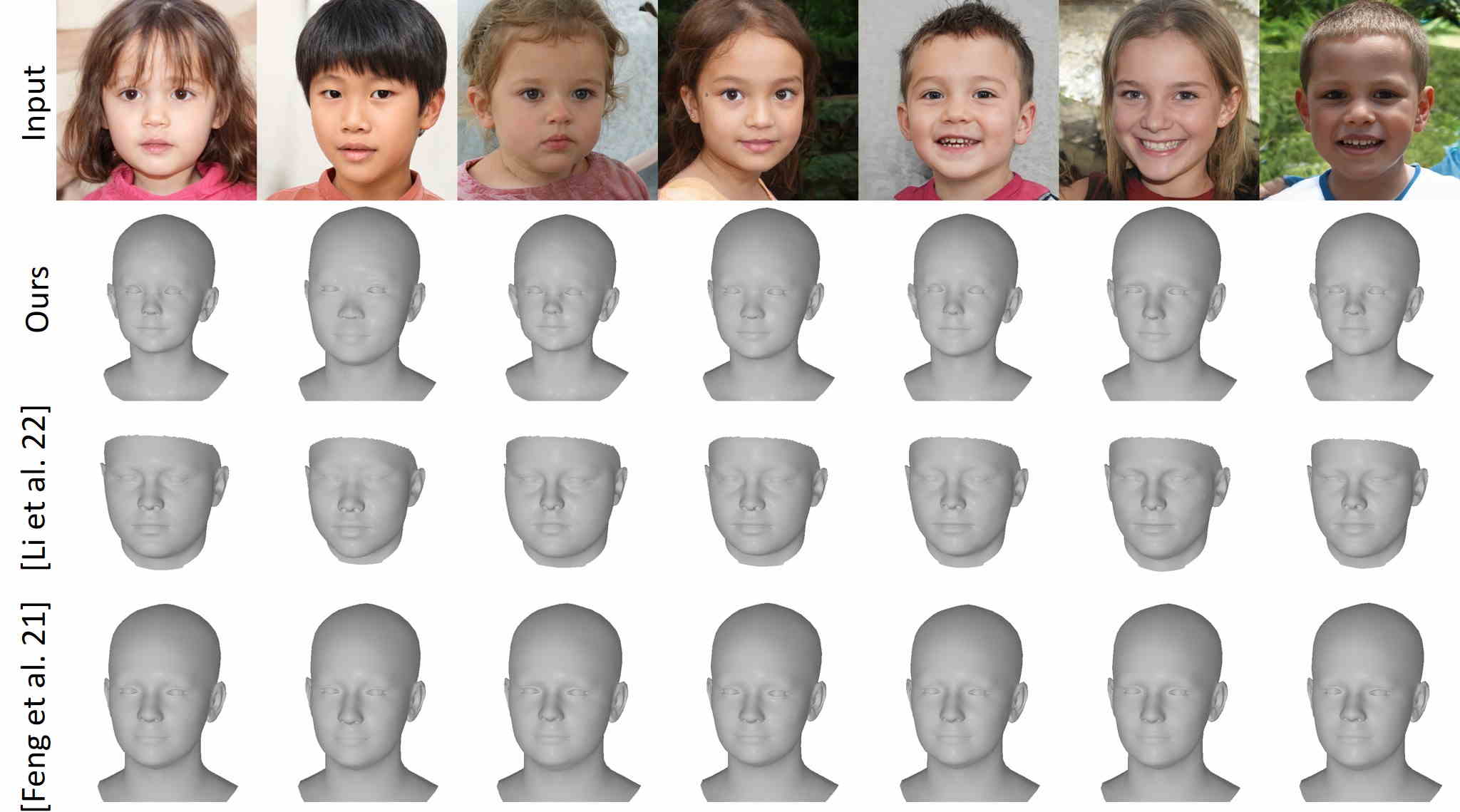}
    \caption{
      Current methods are not predicting metrical faces, which becomes visible when displaying them in a metrical space and not in their image spaces.
      To illustrate we render the prediction of the faces of toddlers in a common metrical space using the same projection.
      State-of-the-art approaches trained in a self-supervised fashion like DECA~\cite{deca} or weakly-supervised like FOCUS~\cite{focus} scale the face of an adult to fit the observation in the image space, thus, the prediction in 3D is non-metrical.
      In contrast, our reconstruction method is able to recover the physiognomy of the toddlers.
      Input images are generated by StyleGan2 \cite{Karras2019stylegan2}.
    }
  \label{fig:kids_shape}
\end{figure}

\medskip \noindent
\textbf{Qualitative Results.}
In \Cref{fig:invariance_qualitative}, we show qualitative results to analyze the stability of the face shape prediction of a subject across different expressions, head rotation, occlusions, or perspective distortion.
As can be seen, our method is more persistent compared to others, especially, in comparison to Deng et al. \cite{deng2019accurate} where shape predictions vary the most.
\Cref{fig:kids_shape} depicts the challenging scenario of reconstructing toddlers from single images.
Instead of predicting a small face for a child, the state of the art methods are predicting faces of adults.
In contrast, MICA predicts the shape of a child with a correct scale.
In \Cref{fig:qualitative_voxceleb} reconstructions for randomly sampled identities from the VoxCeleb2 \cite{voxceleb2} dataset are shown.
Some of the baselines, especially, RingNet~\cite{RingNet:CVPR:2019}, exhibits strong bias towards the mean human face.
In contrast, our method is able to not only predict better overall shape but also to reconstruct challenging regions like nose or chin, even though the training dataset contains a much smaller identity and ethnicity pool.
Note that while the reconstructions of the baseline methods look good under the projection, they are not metric as shown in Tables~\ref{tab:now_challenge} and \ref{tab:feng_challenge}.
%
%

\subsection{Limitations}
Our method is not designed to predict shape and expressions in one forward pass, instead, we reconstruct the expression separately using an optimization-based tracking method.
However, this optimization-based tracking leads to temporally coherent results, as can be seen in the suppl. video.
In contrast to DECA~\cite{deca} or Deng et al.~\cite{deng2019accurate}, the focus of our method is the reconstruction of a metrical 3D model, reconstructing high-frequent detail on top of our prediction is an interesting future direction.
Our method fails, when the used face detector~\cite{retinaface} does not recognize a face in the input.

\begin{figure}[ht!]
    \centering
    \includegraphics[width=0.9\linewidth]{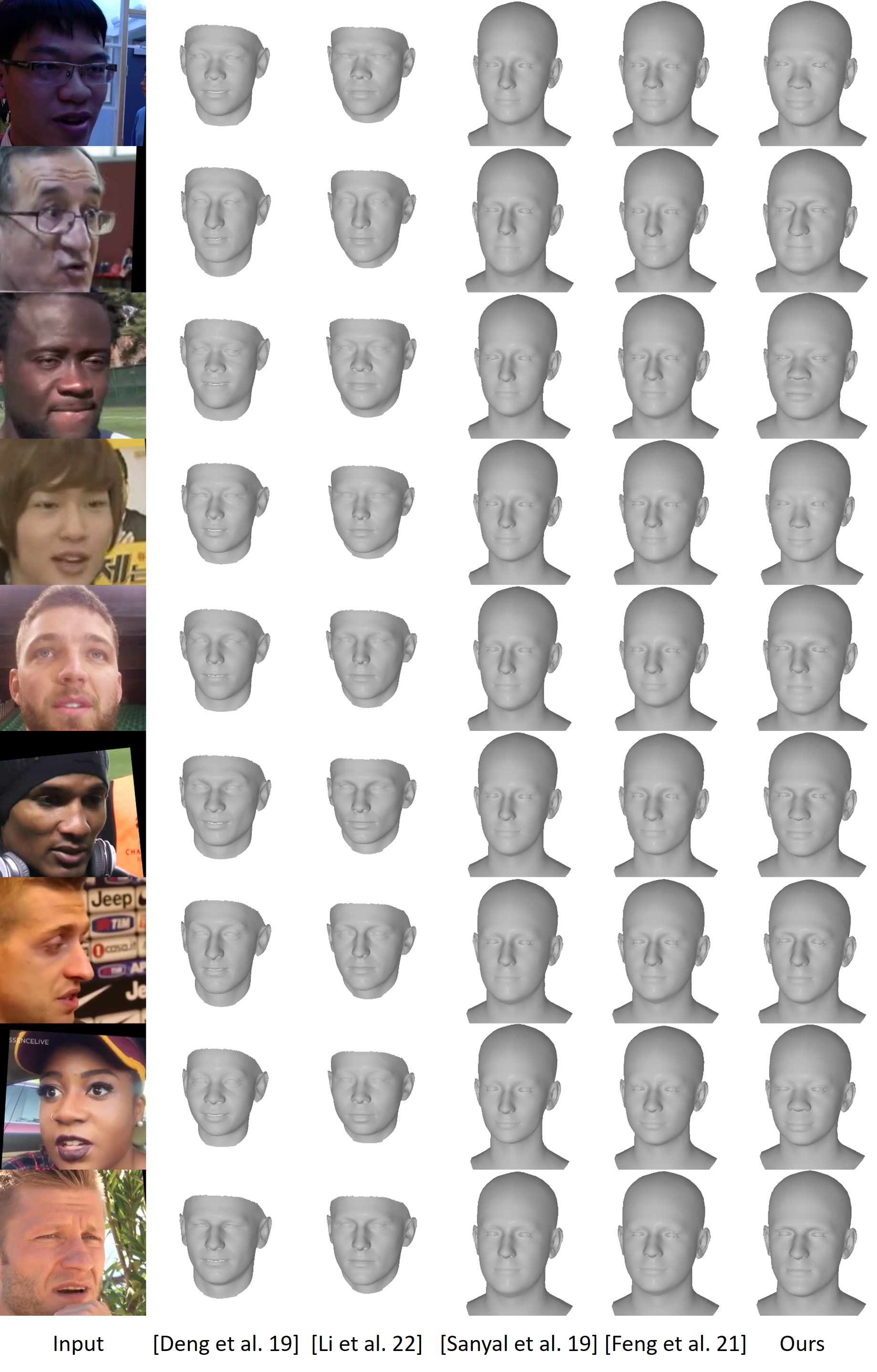}
    \vspace{-0.2cm}
    \caption{Qualitative comparison on randomly sampled images from the VoxCeleb2 \cite{voxceleb2} dataset.
    Our method is able to capture face shape with intricate details like nose and chin, while being metrical plausible (c.f., Tables \ref{tab:now_challenge} and \ref{tab:feng_challenge}).
    }
    \label{fig:qualitative_voxceleb}
\end{figure}

\section{Discussion \& Conclusion}
\label{sec:discussion}
A metrical reconstruction is key for any application that requires the measurement of distances and dimensions.
It is essential for the composition of reconstructed humans and scenes where objects of known size are in, thus, it is especially important for virtual reality and augmented reality applications.
However, we show that recent methods and evaluation schemes are not designed for this task.
While the established benchmarks report numbers in millimeters, they are computed with an optimal scale to align the prediction and the reference.
We strongly argue against this practice, since it is misleading and the errors are not absolute metrical measurements.
To this end, we propose a simple, yet fundamental adjustment of the benchmarks to enable metrical evaluations.
Specifically, we remove the optimal scaling, and only allow rigid alignment of the prediction with the reference shape.
As a stepping stone towards metrical reconstructions, we unified existing small- and medium-scale datasets of paired 2D/3D data.
This allows us to establish 3D supervised losses in our novel shape prediction framework.
While our data collection is still comparably small (around 2k identities), we designed \model that uses features from a face recognition network pretrained on a large-scale 2D image dataset to generalize to in-the-wild image data.
We validated our approach in several experiments and show state-of-the-art results on our newly introduced metrical benchmarks as well as on the established scale-invariant benchmarks.
We hope that this work inspires researchers to concentrate on metrical face reconstruction.

{\small
\paragraph{Acknowledgement.}
We thank Haiwen Feng for support with NoW and Stirling evaluations, and Chunlu Li for providing FOCUS results. 
The authors thank the International Max Planck Research School for Intelligent Systems (IMPRS-IS) for supporting Wojciech Zielonka.

\paragraph{Disclosure.}
While TB is part-time employee of Amazon, his research was performed solely at, and funded solely by MPI.
JT is supported by Microsoft Research gift funds.
}

\pagestyle{headings}
\mainmatter
\def\ECCVSubNumber{7474}  
\title{Towards Metrical Reconstruction \\ of Human Faces \\ --Supplemental Document--}

\titlerunning{Towards Metrical Reconstruction of Human Faces}
\author{Wojciech Zielonka \and
Timo Bolkart \and
Justus Thies}
\authorrunning{W. Zielonka et al.}
\institute{Max Planck Institute for Intelligent Systems, Tübingen}

\maketitle

\begin{figure}
    \centering
    \vspace{-1cm}
    \subfloat[NoW Challenge (non-metrical)]
    {{\includegraphics[width=0.45\linewidth]{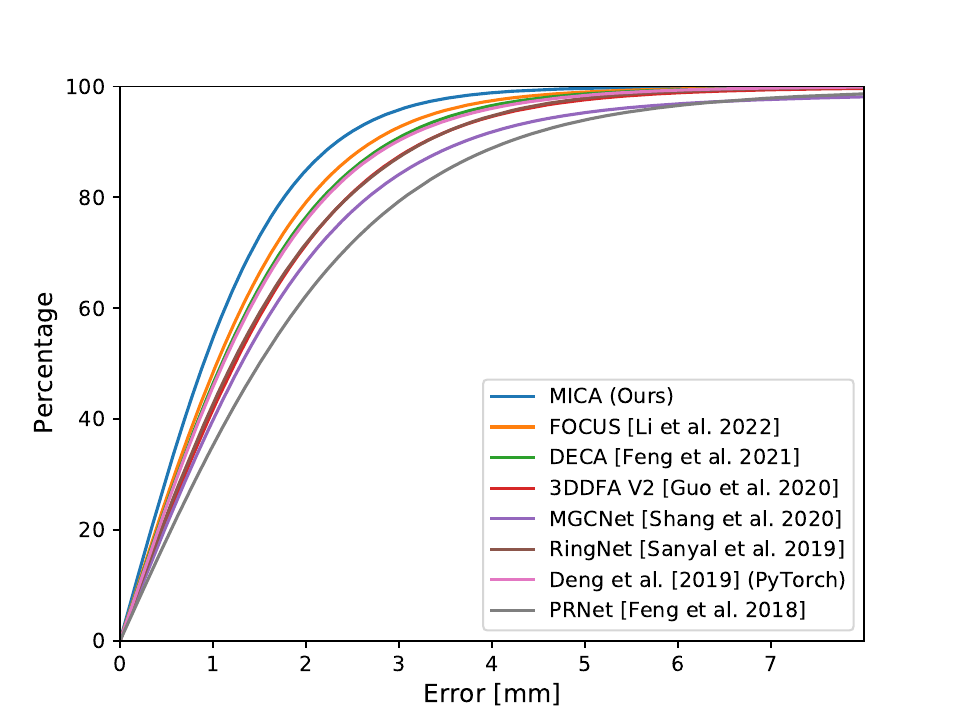} }}%
    \hfill
    \subfloat[NoW-Metric Challenge]{{\includegraphics[width=0.45\linewidth]{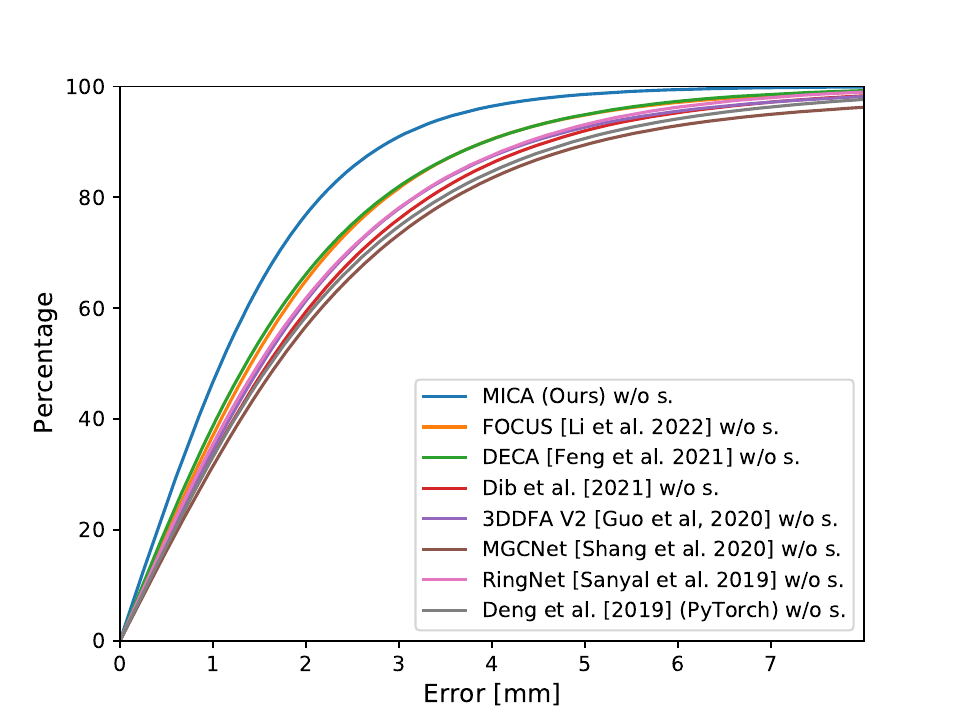} }}%

    \caption[Face Tracking Evaluation]{Cumulative plots for the (a) NoW~\cite{RingNet:CVPR:2019} and (b) NoW-Metric (w/o scale) challenges. We refer to the main paper for the detailed statistics.}
    \label{fig:now_cumulative}
    \vspace{-1cm}
\end{figure}

\begin{abstract}
    In this supplemental document, we demonstrate the robustness of our proposed method in additional qualitative and quantitative experiments.
    The cumulative error plots from the NoW challenge presented in the main paper are also included in this document.
    Moreover, we present a justification of our architecture selection which is tailored for our unified dataset.
    Further, we discuss an alternative model-free estimation approach that does not rely on a 3DMM decoder and can be learned solely on our unified data.
\end{abstract}


\section{Additional Results}
\label{appendix:ablation}
Our 3DMM-based shape estimation method presented in the main paper has two key components, (1) the encoder based on a face recognition network with a mapping network and (2) the 3DMM-based geometry decoder.
The difference between our and the state-of-the-art methods w.r.t. their reconstruction quality gets well visible in the cumulative error plots in \Cref{fig:now_cumulative}.
Moreover, \Cref{fig:qualitative_voxceleb} depicts side views of the reconstructions, which gives a better look at the shape quality.
In this section, we present several ablation studies w.r.t. those modules and the used training data. All experiments were done with the same optimizer and hyper-parameter configuration as the main method except where stated otherwise. The Stirling dataset was excluded from all the ablation experiments.

\begin{figure}[H]
    \centering
    \includegraphics[width=0.9\linewidth]{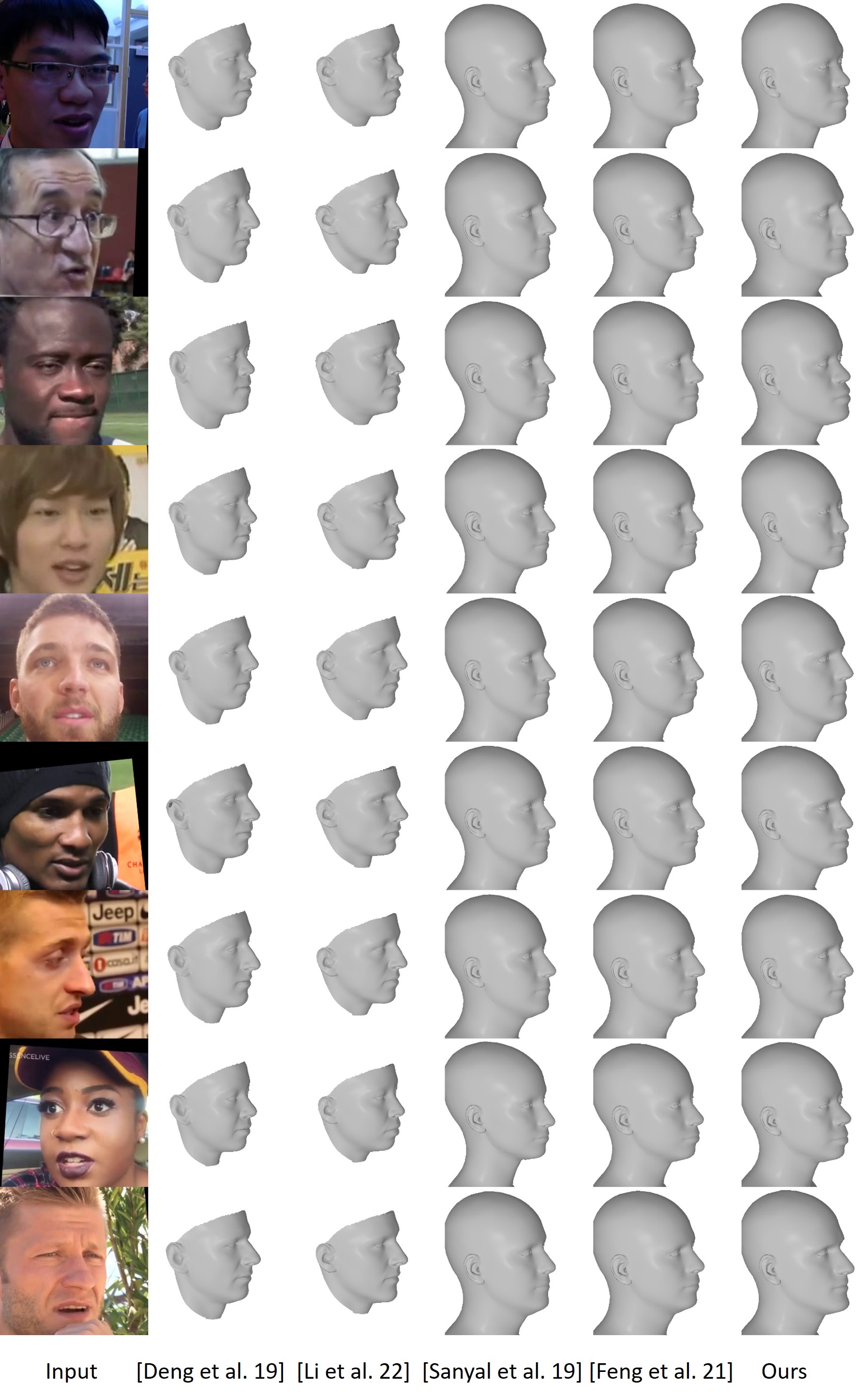}
    \vspace{-0.2cm}
    \caption{Qualitative comparison on randomly sampled images from the VoxCeleb2 \cite{voxceleb2} dataset for side views.
    }
    \label{fig:qualitative_voxceleb}
\end{figure}

\paragraph{Encoder Ablation Studies.}
Exploiting generalized facial features from a face recognition network is a key component of our method to predict geometry from in-the-wild 2D data. However, completely refining the latent space of the face recognition network is not possible with our medium-size dataset, thus, we can only retrain selected layers to maintain generalizability.
In \Cref{tab:face_recon_ablation}, we compare the performance of the two face recognition methods ArcFace \cite{arcface} and FaceNet \cite{facenet}.
Overall, the pretrained ArcFace outperforms the pretrained FaceNet in terms of reconstruction quality in our shape estimation architecture.
To further improve the results of ArcFace, we refine the last ResNet layer of ArcFace.
Similarly, we conducted experiments on fine-tuning DECA~\cite{deca} using our medium-scale dataset and our reconstruction loss based on an $\ell_1$ error metric. We trained the network on the same datasets like ArcFace for around 500 epochs.
The fine-tuning of partial layers or entire pipeline leads to huge overfitting of the training data with significantly worse reconstructions on the test dataset (see \Cref{tab:face_recon_ablation}).
In contrast, the partial fine-tuning of ArcFace in our approach gives the lowest mean reconstruction error of $1.35$mm.
It shows that we can effectively use the generalized features from the ArcFace network for the task of metrical face reconstruction.
\begin{table}
    \caption{Ablation study w.r.t. our face encoding network based on Stirling  dataset~\cite{stirling} using our metrical evaluation scheme.
    As a comparison, we also show the results for DECA~\cite{deca} fine-tuned on our dataset. The respective ResNet 
    \cite{resnet} networks were refined in different configurations; $\{L3, L4\}$ denotes the set of selected trainable layers from $\{L1, ..., L4\}$. Each layer is composed of several ResNet blocks, specifically, ArcFace uses $\{3, 13, 30, 3\}$ and DECA $\{3, 4, 6, 3\}$ ResNet blocks for the respective layers.
    }
    \centering
    \medskip
    \begin{tabular} { l|c|c|c|c|c|c }
    \toprule 
    \multirow{2}{*}{\textbf{Encoder}}
    & \multicolumn{2}{c|}{Median}
    & \multicolumn{2}{c|}{Mean (mm)}
    & \multicolumn{2}{c}{Std} \\
    \cline{2-7}
    & LQ & HQ & LQ & HQ & LQ & HQ \\
    \toprule 
      DECA ~\cite{deca}  (frozen) & $1.32$ & $1.22$ & $1.71$ & $1.58$ & $1.54$ & $1.42$\\ 
      DECA ~\cite{deca}  (fully trainable) & 1.54 & 1.42 & 1.96 & 1.82 & 1.71 & 1.61 \\ 
      DECA ~\cite{deca}  ($L3-L4$ trainable) & 1.55 & 1.43 & 1.97 & 1.83 & 1.71 & 1.62 \\ 
      DECA ~\cite{deca}  ($L4$ trainable) & 1.55 & 1.49 & 1.97 & 1.83 & 1.71 & 1.61 \\ 
    \hline
      Ours -- FaceNet~\cite{facenet} (frozen) & 1.37 & 1.29 & 1.75 & 1.65 & 1.56 & 1.47 \\ 
      Ours -- ArcFace~\cite{arcface} (frozen) & 1.25 & 1.18 & 1.60 & 1.52 & 1.43 & 1.37 \\ 
      Ours -- ArcFace~\cite{arcface} (fully trainable) & 1.18 & 1.11 & 1.52 & 1.42 & 1.38 & 1.27 \\ 
      Ours -- ArcFace~\cite{arcface} ($L2-L3-L4$ trainable) & 1.22 & 1.12 & 1.56 & 1.43 & 1.39 & 1.27 \\ 
      Ours -- ArcFace~\cite{arcface} ($L3-L4$ trainable) & 1.17 & 1.10 & 1.51 & 1.40 & 1.37 & 1.25 \\ 
      Ours -- ArcFace~\cite{arcface} ($L4$ trainable) & 1.15 & 1.06 & 1.46  & 1.35 & 1.30 & 1.20 \\ 
    \bottomrule
    \end{tabular}
    \label{tab:face_recon_ablation}
\end{table}

\newpage
\paragraph{Decoder Ablation Studies.}
The decoder is defined by the 3DMM FLAME~\cite{flame}.
For our experiments in the main paper, we used $300$ eigenvectors of the PCA basis.
In \Cref{tab:num_shape_params_ablation}, we present an ablation study on the number of used eigenvectors (i.e., the size of the latent geometry code $\boldsymbol{z}$).
As can be seen, exploiting the full linear space of FLAME leads to the best performance.

\begin{table}[H]
    \caption{Evaluation of the influence of the number of principle components (PCs) used for the shape decoder (Stirling dataset~\cite{stirling} with NoW protocol (metrical)).}
    \centering
    \medskip
    \begin{tabular} { l|c|c|c|c|c|c }
    \toprule 
    \multirow{2}{*}{\textbf{Decoder - \#PC}}
    & \multicolumn{2}{c|}{Median}
    & \multicolumn{2}{c|}{Mean (mm)}
    & \multicolumn{2}{c}{Std} \\
    \cline{2-7}
    & LQ & HQ & LQ & HQ & LQ & HQ \\
    \toprule 
      50 & 1.19 & 1.12 & 1.50 & 1.41 & 1.33 & 1.23 \\ 
      100 & 1.15 & 1.10 & 1.45 & 1.39 & 1.28 & 1.22 \\ 
      200 & 1.15 & 1.06 & 1.47 & 1.36 & 1.31 & 1.20 \\ 
      300 & 1.15 & 1.06 & 1.46 & 1.35 & 1.30 & 1.20 \\ 
    \bottomrule
    \end{tabular}
    \label{tab:num_shape_params_ablation}
\end{table}

\paragraph{Dataset Ablation Studies.}
As described in the main paper, we used several datasets to construct our training set.
We perform a leave-one-out analysis in \Cref{tab:now_datasets_ablation} on the Stirling dataset.
The LYHM dataset contains 1211 subjects and, thus, has the most significant influence on the training.

\medskip
\noindent
In addition to the datasets listed in the main paper, we also processed FaceScape~\cite{zhu2021facescape,yang2020facescape}.
While FaceScape is a large-scale dataset, it has been recorded within an uncalibrated setup, thus, being in a none metrical scale which is introducing a bias in our prediction.

\begin{table}[H]
    \caption{To analyze the contribution of a single dataset, we perform a leave-one-out analysis.
    We report the reconstruction quality for images from Stirling dataset~\cite{stirling} (where we exclude Stirling from training).
    As can be seen, LYHM~\cite{lyhm} has the highest influence on the reconstruction quality, leaving it out leads to an increase of the mean error for HQ images from $1.35$mm to $1.43$mm and for LQ images from $1.46$mm to $1.51$mm on the Stirling dataset \cite{stirling}.}
    \medskip
    \centering
    \begin{tabular} { l|c|c|c|c|c|c }
    \toprule 
    \multirow{2}{*}{\textbf{Dataset}}
    & \multicolumn{2}{c|}{Median}
    & \multicolumn{2}{c|}{Mean (mm)}
    & \multicolumn{2}{c}{Std} \\
    \cline{2-7}
    & LQ & HQ & LQ & HQ & LQ & HQ \\
    \toprule 
      w/o LYHM~\cite{lyhm} & 1.18 & 1.12 & 1.51 & 1.43 & 1.35 & 1.27 \\
      w/o FRGC~\cite{bu3dfe} & 1.15 & 1.06 & 1.47 & 1.36 & 1.33 & 1.23 \\ 
      w/o BP4D+~\cite{bu3dfe} & 1.14 & 1.09 & 1.46 & 1.38 & 1.30 & 1.21 \\ 
      w/o BU3DFE~\cite{bu3dfe} & 1.14 & 1.08 & 1.45 & 1.37 & 1.29 & 1.21 \\ 
      w/o D3DFACS~\cite{D3DFACS} & 1.13 & 1.06 & 1.43 & 1.35 & 1.28 & 1.19 \\ 
      w/o Face Warehouse~\cite{D3DFACS} & 1.13 & 1.07 & 1.43 & 1.36 & 1.27 & 1.20 \\ 
      All & 1.15 & 1.06 & 1.46 & 1.35 & 1.30 & 1.20 \\
    \bottomrule
    \end{tabular}
    \label{tab:now_datasets_ablation}
\end{table}


\section{Studies on the Facial Expression Tracking}
\label{appendix:tracking}
\begin{figure}[ht]
    \centering
    \includegraphics[width=0.75\linewidth]{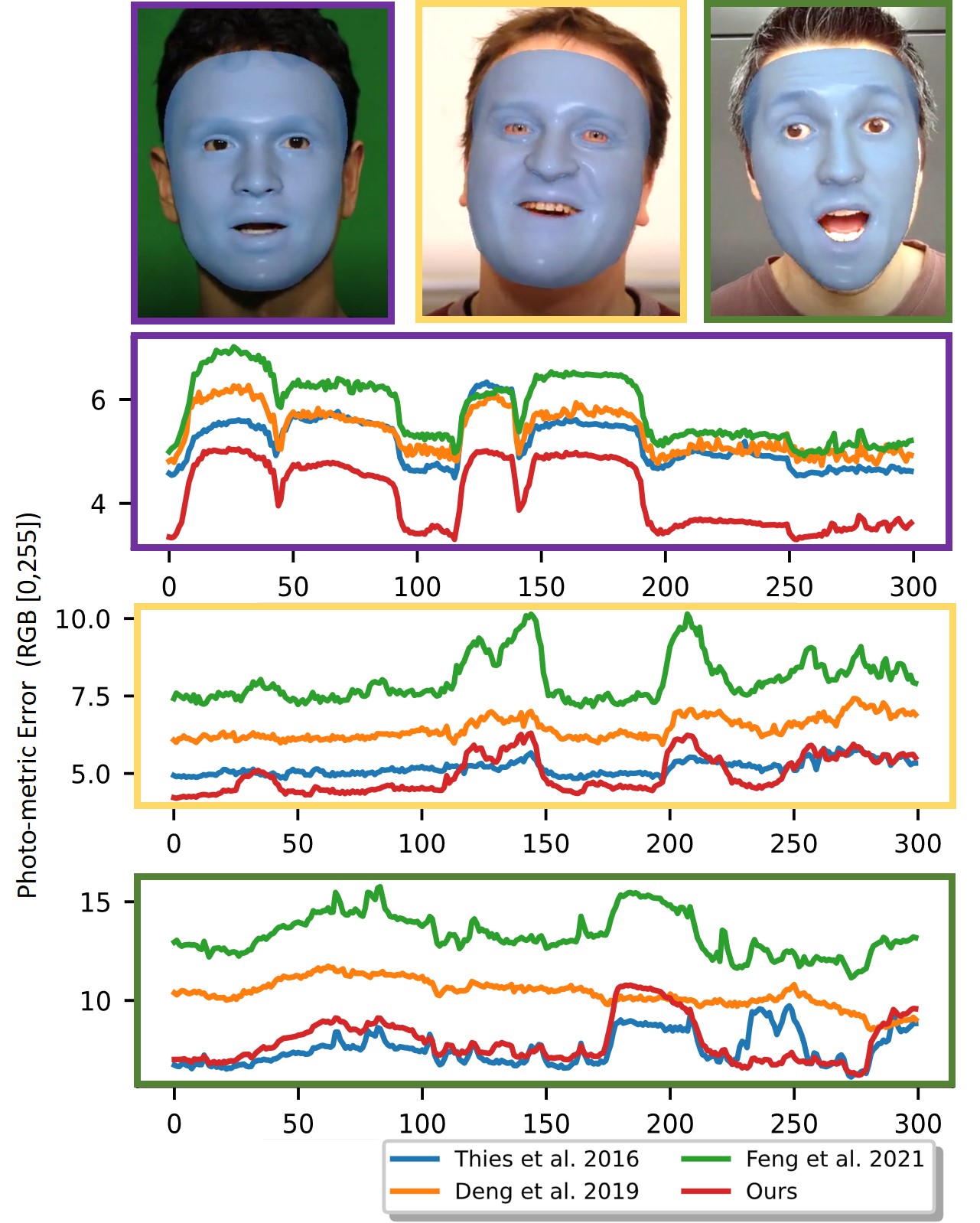}
    \caption[Face Tracking Evaluation]{Photo-metric error on the three sequences from Garrido et al.~\cite{GVWT13} shown in the supplemental video.The photo-metric error is computed based on an RMSE error metric assuming RGB in  $[0,255]^3$.}
    \label{fig:tracking_error_color}
\end{figure}

Our metrical face shape prediction can be used to initialize facial expression tracking.
In contrast to methods like \cite{deca}, our method uses a perspective camera model, which allows us to predict a depth.
In \Cref{fig:tracking_error_color}, we show a sample sequence from \cite{VWBS12} with an according depth and photo-metric error plot.
\begin{figure}[ht]
    \centering
    \includegraphics[width=0.75\linewidth]{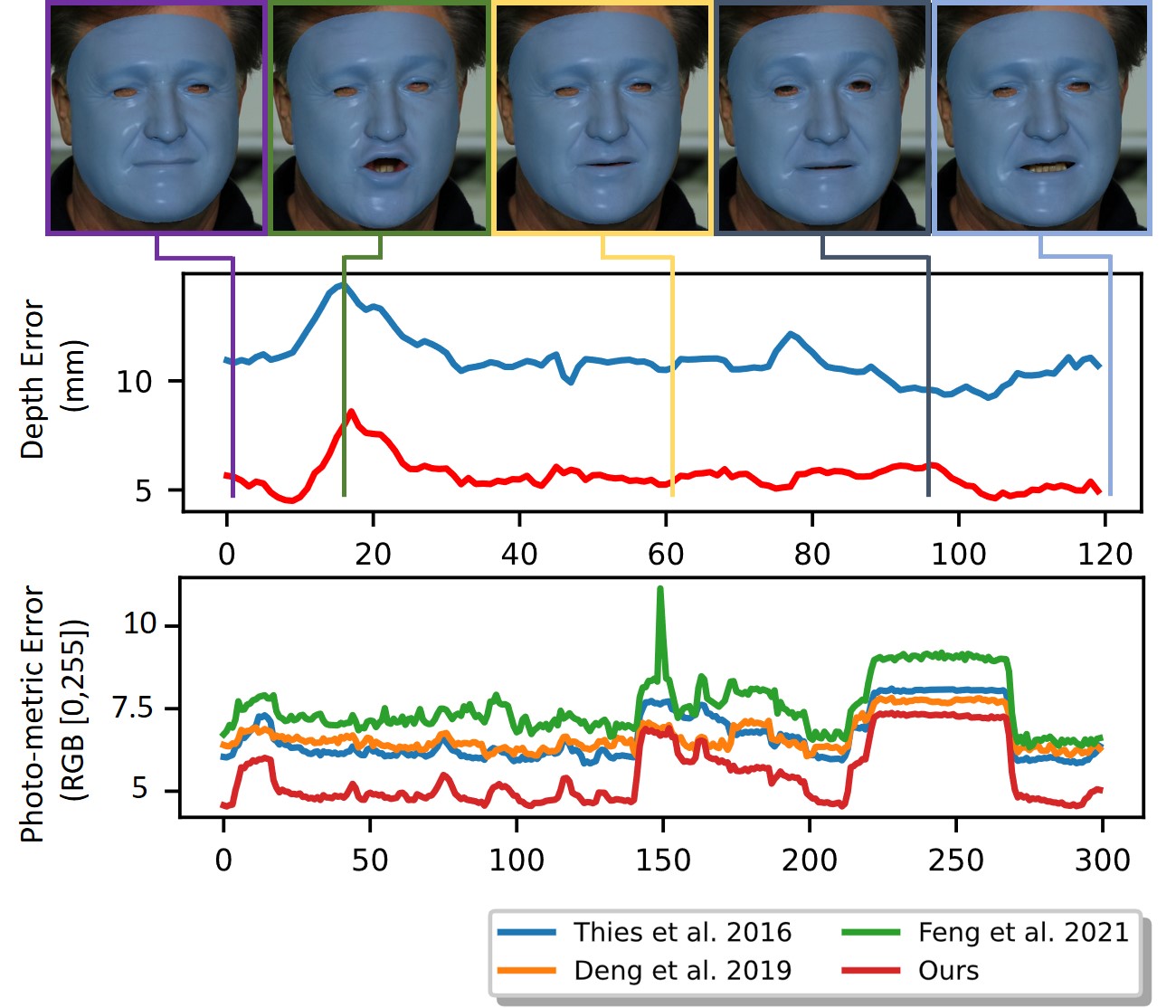}
    \caption[Face Tracking Evaluation]{Evaluation of the tracking error on the 'Volker' sequence of \cite{VWBS12}. The RMSE depth error is computed based on the reference depth maps which have been reconstructed using a stereo system. The photo-metric error is computed based on an RMSE error metric assuming RGB in  $[0,255]^3$.
    }
    \label{fig:tracking_error_depth}
\end{figure}

As can be seen, our method results in the lowest photo-metric error in terms of a masked RMSE metric on the colors.
The error plots shown in \Cref{fig:tracking_error_depth} contain the metric reconstruction error of the depth (RMSE).
It is based on the reference depth information of the sequence, which has been reconstructed from a passive stereo system.
We also evaluate the dense photometric error (RMSE), which can be computed for \cite{deca,deng2019accurate} too.
In comparison to the method Face2Face~\cite{face2face} which also uses a perspective camera model ($11.0$mm mean RMSE depth error), our metrical face shape estimation improves the tracking quality significantly ($5.7$mm mean RMSE).
In the supplemental video, we show several tracking results which demonstrate that our proposed technique is temporally stable.

%

\newpage
\section{Model-free Decoder}
\label{appendix:siren}

\begin{figure}[H]
    \centering
    \includegraphics[width=0.8\textwidth]{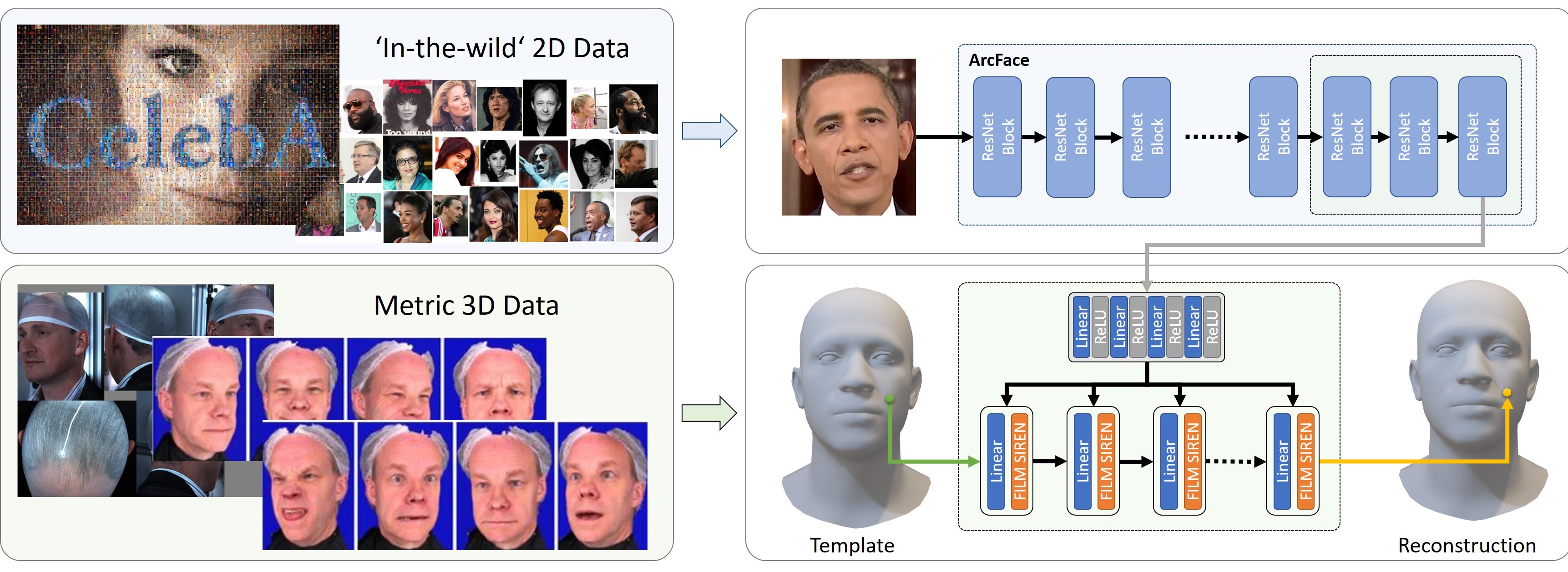}
    \caption{
        Overview of a model-free decoder. The model-free decoder is based on a Siren architecture~\cite{siren} using FiLM conditionings~\cite{piGAN}.
        In contrast to the FLAME-based decoder, this model-free decoder can be trained in conjunction to the encoder only based on the dataset with the paired 2D/3D data which is smaller than the dataset of 3D scans used for constructing the FLAME model.
    }
    \label{fig:pipeline_siren}
\end{figure}

Inspired by pi-GAN~\cite{piGAN} and Dynamic Surface Function Networks~\cite{burov2021dsfn}, we also evaluated a coordinated-based multi layer perceptron (MLP) with sinusoidal activation functions (SIREN~\cite{siren}) to represent the geometry of a face (see \Cref{fig:pipeline_siren}).
This architecture can be trained solely on the data of our unified dataset without requiring any 3DMM model.
The network and its sinusoidal activation functions are controlled by a mapping network $\mathcal{M}'$ to represent different faces.
The mapping network takes the identity code $z$ as input and predicts the frequencies and phase shifts of the sinusoidal activation layers.
The SIREN network $\mathcal{S}$ is evaluated at the FLAME~\cite{flame} template mesh vertices $\boldsymbol{A} \in \mathbb{R}^{3N}$ and $N=5023$ to leverage the correspondences of the 3D training data.
\begin{equation*}
    \mathcal{G}_{SIREN}(\boldsymbol{z}) = \mathcal{S}(\boldsymbol{A} ~|~ \mathcal{M}'(\boldsymbol{z})).
\end{equation*}
Since this model does not rely on the PCA basis of the FLAME model, it can predict meshes outside the FLAME face space.
In comparison to the 3DMM-based model presented in the main paper, this model-free approach performs on par on the different benchmarks (see \Cref{tab:feng_now_challenge_siren}).
A benefit of the model-free decoder is that it can be trained solely on our dataset of paired 2D/3D data which is significantly smaller than the dataset of 3D scans used for the construction of the FLAME model (2.3k (our dataset) versus 4k subjects used for FLAME).

A drawback of this Siren-based approach is its runtime and complexity (3DMM only has a single linear layer for representing shape variations).
The used SIREN network is a more compact representation using $8$ hidden layers and $256$ feature size with total $1976327$ parameters, while the 3DMM has a linear layer with $(300+1)*5023*3 = 4535769$ parameters.

\begin{table}
        \caption{
        Quantitative evaluation of the face shape estimation using Stirling dataset~\cite{stirling} and NoW protocol (metrical).
    }
    \centering
    \medskip
    \resizebox{\linewidth}{!}{
    \begin{tabular}{l| *{11}{c|}c }
    \toprule 
    & \multicolumn{6}{c|}{Non-Metrical} 
    & \multicolumn{6}{c}{Metrical (mm)}\\
    \cline{2-13}
    \textbf{Stirling (NoW Protocol)} & \multicolumn{2}{c|}{Median}
    & \multicolumn{2}{c|}{Mean}
    & \multicolumn{2}{c|}{Std}
    & \multicolumn{2}{c|}{Median}                
    & \multicolumn{2}{c|}{Mean}          
    & \multicolumn{2}{c}{Std}          
    \\
    \cline{2-13}
      & LQ & HQ & LQ & HQ & LQ & HQ & LQ & HQ & LQ & HQ & LQ & HQ \\
    \toprule
    Deng et al.~\cite{deng2019accurate} (PyTorch)
    & $1.12$ & $0.99$ & $1.44$ & $1.27$ & $1.31$ & $1.15$
    & $1.47$ & $1.31$ & $1.93$ & $1.71$ & $1.77$ & $1.57$
    \\
    DECA~\cite{deca}
    & $1.09$ & $1.03$ & $1.39$ & $1.32$ & $1.26$ & $1.18$
    & $1.32$ & $1.22$ & $1.71$ & $1.58$ & $1.54$ & $1.42$
    \\
    Ours (SIREN)
    & $1.01$ & $0.94$ & $1.28$ & $1.19$ & $1.15$ & $1.06$
    & $1.20$ & $1.09$ & $1.53$ & $1.39$ & $1.35$ & $1.23$
    \\
    \textbf{Ours (FLAME)}
    & \textbf{0.96} & \textbf{0.92} & \textbf{1.22} & \textbf{1.16} & \textbf{1.11} & \textbf{1.04}
    & \textbf{1.15} & \textbf{1.06} & \textbf{1.46} & \textbf{1.35} & \textbf{1.30} & \textbf{1.20} \\
    \bottomrule
    \end{tabular}
    }

    \label{tab:feng_now_challenge_siren}
\end{table}

\clearpage
\bibliographystyle{splncs04}
\bibliography{main}
\end{document}